\pdfoutput=1
\documentclass[final]{cvpr}

\usepackage[utf8]{inputenc} %
\usepackage{times}
\usepackage{url}            %
\usepackage{booktabs}       %
\usepackage{amsmath}
\usepackage{amssymb}
\usepackage{amsfonts}       %
\usepackage{nicefrac}       %
\usepackage{microtype}      %
\usepackage{graphicx}
\usepackage[usenames, dvipsnames]{xcolor}
\usepackage{multirow}
\usepackage{enumitem}
\usepackage{xspace}
\usepackage{tikz}
\usepackage[capbesideposition=outside,capbesidesep=quad]{floatrow}		%
\usepackage[numbers,sort,compress]{natbib} %

\usepackage[pagebackref=true,breaklinks=true,colorlinks,bookmarks=false]{hyperref}

\newcommand{\boldparagraph}[1]{\vspace{0.1em}\noindent{\bf #1} }

\definecolor{darkgreen}{rgb}{0,0.5,0}
\definecolor{semidarkgreen}{rgb}{0,0.8,0}

\def\eg{\emph{e.g}\onedot}

\def\etal{\emph{et al}\onedot}

\newcommand{\nN}{\mathbb{N}}

\newcommand{\nR}{\mathbb{R}}

\newcommand{\nLoss}{\mathcal{L}}
\newcommand{\nFeatDim}{N}
\newcommand{\nFeatGrid}{g}
\newcommand{\nLFeatGrid}{v}
\newcommand{\nTSDF}{s}
\newcommand{\nOcc}{o}

\newcommand{\nDepth}{D}
\newcommand{\nQueryPoint}{p}

\begin{document}

\title{NeuralFusion: Online Depth Fusion in Latent Space}

\author{Silvan Weder$^1$
	\qquad
	Johannes L.~Sch\"{o}nberger$^2$
	\qquad
	Marc Pollefeys$^{1,2}$
	\qquad
	Martin R.~Oswald$^1$\\
	$^1$Department of Computer Science, ETH Zurich\\
	$^2$Microsoft Mixed Reality and AI Zurich Lab}

\maketitle
\pagestyle{empty}

\begin{abstract}
We present a novel online depth map fusion approach that learns depth map aggregation in a latent feature space.
While previous fusion methods use an explicit scene representation like signed distance functions (SDFs), we propose a learned feature representation for the fusion. 
The key idea is a separation between the scene representation used for the fusion and the output scene representation, via an additional translator network.
Our neural network architecture consists of two main parts: a depth and feature fusion sub-network, which is followed by a translator sub-network to produce the final surface representation (e.g. TSDF) for visualization or other tasks.
Our approach is an online process, handles high noise levels, and is particularly able to deal with gross outliers common for photometric stereo-based depth maps.
Experiments on real and synthetic data demonstrate improved results compared to the state of the art, especially in challenging scenarios with large amounts of noise and outliers. 
The source code will be made available at \href{https://github.com/weders/NeuralFusion}{https://github.com/weders/NeuralFusion}.
\end{abstract}

\section{Introduction}
Reconstructing the geometry of a scene is a central component of many applications in 3D computer vision. 
Awareness of the surrounding geometry enables robots to navigate, augmented and mixed reality devices to accurately project the information into the user's field of view, and serves as the basis for many 3D scene understanding tasks.

In this paper, we consider online surface reconstruction by fusing a stream of depth maps with known camera calibration.
The fundamental challenge in this task is that depth maps are typically noisy, incomplete, and contain outliers.
Depth map fusion is a key component in many 3D reconstruction methods, like KinectFusion~\cite{Newcombe-et-al-ISMAR-2011}, VoxelHashing~\cite{Niessner-et-all-ACM-2013}, InfiniTAM~\cite{Kaehler-et-al-TVCG-2015}, and many others.
The vast majority of methods builds upon the concept of averaging truncated signed distance functions (TSDFs), as proposed in the pioneering work by Curless and Levoy~\cite{Curless-et-al-SIGGRAPH-1996}.
This approach is so popular due to its simple, highly parallelizable, and real-time capable way of fusing noisy depth maps into a surface. 
However, it has difficulties with handling outliers and thin geometry, which can be mainly attributed to the local integration of depth values in the TSDF volume.

To tackle this fundamental limitation, existing methods use various heuristics to filter outliers in decoupled pre- or post-processing steps.
Such filtering techniques entail the usual trade-off in terms of balancing accuracy against completeness.
Especially in an online fusion system, striking this balance is extremely challenging in the pre-filtering stage, because it is difficult to distinguish between a first surface measurement and an outlier.
Consequently, to achieve complete surface reconstructions, one must use conservative pre-filtering, which in turn requires careful post-filtering of outliers by non-local reasoning on the TSDF volume or the final mesh.
\begin{figure}[t]
	\centering
	\scriptsize
	\setlength{\tabcolsep}{1pt}
	\newcommand{\sz}{4.2cm}
	\begin{tabular}{cccc}
		\includegraphics[height=\sz]{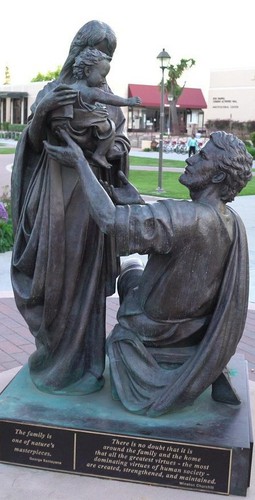} &
		\includegraphics[height=\sz]{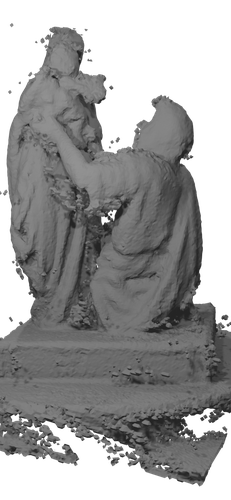} &
		\includegraphics[height=\sz]{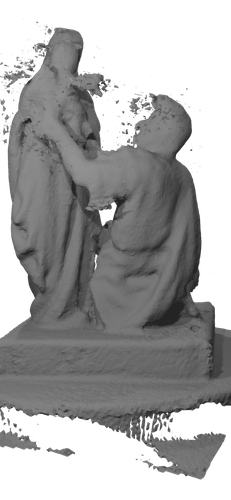} &
		\includegraphics[height=\sz]{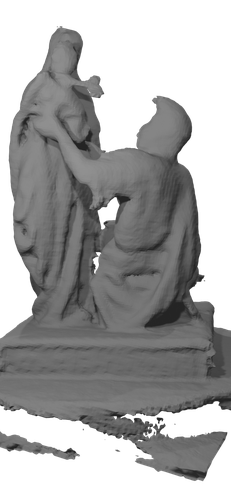} \\
		& 
		TSDF Fusion~\cite{Curless-et-al-SIGGRAPH-1996} & 
		RoutedFusion~\cite{Weder-et-al-CVPR-2020} & 
		Ours\\[-6pt]
	\end{tabular}
	\caption{\textbf{Results of our end-to-end depth fusion} on real-world MVS data~\cite{Knapitsch-et-al-TOG-2017}. Our method learns to separate outliers and true geometry without the need of filtering heuristics.}
	\label{fig:teaser}
	\vspace{2em} %
\end{figure}
This motivates our key idea to use different scene representations for the fusion and final method output, while all prior works perform depth fusion and filtering directly in the output representation.
In contrast, we perform the fusion step in a latent scene representation, implicitly learned to encode features like confidence information or local scene information.
A final translation step simultaneously filters and decodes this learned scene representation into the final output relevant to downstream applications.

\newpage
\noindent
In summary, we make the following \textbf{contributions:}
\begin{itemize}[itemsep=0pt,topsep=3pt,leftmargin=*]
\item We propose a novel, end-to-end trainable network architecture, which separates the scene representations for depth fusion and the final output into two different modules.
\item The proposed latent representation yields more accurate and complete fusion results for larger feature dimensions allowing to balance accuracy against resource demands. 
\item Our network architecture allows for end-to-end learnable outlier filtering within a translation step that significantly improves outlier handling.
\item Although fully trainable, our approach still only performs very localized updates to the global map which maintains the online capability of the overall approach.
\end{itemize}

\section{Related Work}

\boldparagraph{Representation Learning for 3D Reconstruction.}
Many works proposed learning-based algorithms using implicit shape representations.
Ladicky~\etal~\cite{Ladicky-et-al-ICCV-2017} learn to regress a mesh from a point cloud using a random forest.
The concurrent works OccupancyNetworks~\cite{Mescheder-et-al-CVPR-2019}, DeepSDF~\cite{Park-et-al-CVPR-2019}, and IM-NET~\cite{Chen-et-al-CVPR-2019} encode the object surface as the decision boundary of a trained classification or regression network. 
These methods only use a single feature vector to encode an entire scene or object within a unit cube and are, unlike our approach, not suitable for large-scale scenes.
This was improved in \cite{Chabra-et-al-ECCV-2020,Chibane-et-al-CVPR-2020,Jiang-et-al-CVPR-2020,Peng-et-al-ECCV-2020} which use multiple features to encode parts of the scene.
\cite{Kar-et-al-NIPS-2017, Huang-et-al-ECCV-2018, Saito-et-al-ICCV-2019,Saito-et-al-CVPR-2020, Choy-et-al-ECCV-2016} extract image features and learn scene occupancy labels from single or multi-view images.
DISN~\cite{Xu-et-al-NIPS-2019} works in a similar fashion but regresses SDF values.
Chiyu~\etal~\cite{Chiyu-et-al-CVPR-2020} propose a local implicit grid representation for 3D scenes. However, they encode the scene through direct optimization of the latent representation through a pre-trained neural network.
More recently, Scene Representation Networks~\cite{Sitzmann-et-al-NIPS-2019} learn to reconstruct novel views from a single RGB image.
Liu~\etal~\cite{Liu-et-al-NIPS-2019} learn implicit 3D shapes from 2D images in a self-supervised manner.
These works also operate within a unit cube and are difficult to scale to larger scenes.
DeepVoxels~\cite{Sitzmann-et-al-CVPR-2019} encodes visual information in a feature grid with a neural network to generate novel, high-quality views onto an observed object.
Nevertheless, the work is not directly applicable to a 3D reconstruction task.
Our method combines the concept of learned scene representations with data fusion in a learned latent space.
Recently, several works proposed a more local scene representation~\cite{Genova-et-al-2019,Genova-et-al-ICCV-2019,Badki-et-al-2020} that allows larger scale scenes and multiple objects.
Further, \cite{Liu-et-al-CVPR-2020,Niemeyer-et-al-CVPR-2020} learn implicit representations with 2D supervision via differentiable rendering.
Overall, none of all mentioned works consider online updates of the shape representation as new information becomes available and adding such functionality is by no means straightforward.

\boldparagraph{Classic Online Depth Fusion Approaches.}
The majority of depth map fusion approaches are built upon the seminal ``TSDF Fusion'' work by Curless and Levoy~\cite{Curless-et-al-SIGGRAPH-1996}, which fuses depth maps using an \textit{implicit representation} by averaging TSDFs on a dense voxel grid.
This approach especially became popular with the wide availability of low-cost depth sensors like the Kinect and has led to works like KinectFusion~\cite{Newcombe-et-al-ISMAR-2011} and related works like sparse-sequence fusion~\cite{Yang-et-al-TVGC-2018}, BundleFusion~\cite{Dai-et-et-al-SIGGRAPH-2017}, or variants on sparse grids like VoxelHashing~\cite{Niessner-et-all-ACM-2013}, InfiniTAM~\cite{Priscariu-et-al-CoRR-2017}, Voxgraph~\cite{Reijgwart-et-al-RAL-2020}, octree-based approaches~\cite{Fuhrmann-Goesele-TOG-2011,Steinbruecker-et-al-ICCV-2013,Marniok-et-al-GCPR-2017}, or hierarchical hashing~\cite{Kaehler-et-al-RAL-2016}.
However, the output of these methods usually contains typical noise artifacts, such as surface thickening and outlier blobs.
Another line of works uses a \textit{surfel-based representation} and directly fuses depth values into a sparse point cloud~\cite{Stueckler-Behnke-JVCIR-2014,Keller-et-al-3DV-2013,Lefloch-et-al-FUSION-2015,Whelan-et-al-IJRR-2016,McCormac-et-al-ICRA-2017,Schoeps-et-al-TPAMI-2019}.
This representation perfectly adapts to inhomogeneous sampling densities, requires low memory storage, but also lacks connectivity and topological surface information.
An overview of depth fusion approaches is given in~\cite{Zollhoefer-et-al-CGF-STAR-2018}.

\boldparagraph{Classic Global Depth Fusion Approaches.} 
While online approaches only process one depth map at a time, global approaches use all information at once and typically apply additional smoothness priors like total variation~\cite{Zach-et-al-ICCV-2007,Kolev-et-al-IJCV-2009}, its variants including semantic information~\cite{Haene-et-al-CVPR-2013,Cherabier-et-al-3DV-2016,Haene-et-al-TPAMI-2017,Savinov-et-al-CVPR-2015,Savinov-et-al-CVPR-2016}, or refine surface details using color information~\cite{Zollhoefer-et-al-ACM-2015}.
Consequently, their high compute and memory requirements prevent their application in online scenarios unlike ours.

\boldparagraph{Learned Global Depth Fusion Approaches.}
Octnet~\cite{Riegler-et-al-CVPR-2017} and its follow-up OctnetFusion~\cite{Riegler-et-al-3DV-2017} fuse depth maps using TSDF fusion into an octree and then post-processes the fused geometry using machine learning.
RayNet~\cite{Paschalidou-et-al-CVPR-2018} uses a learned Markov random field and a view-invariant feature representation to model view dependencies.
SurfaceNet~\cite{Ji-et-al-ICCV-2017} jointly estimates multi-view stereo depth maps and the fused geometry, but requires to store a volumetric grid for each input depth map.
3DMV~\cite{Dai-et-al-ECCV-2018} optimizes shape and semantics of a pre-fused TSDF scene of given 2D view information.
Contrary to these methods, we learn online depth fusion and can process an arbitrary number of input views.

\boldparagraph{Learned Online Depth Fusion Approaches.}
In the context of simultaneous localization and mapping, CodeSLAM~\cite{Bloesch-et-al-CVPR-2018}, SceneCode~\cite{Zhi-et-al-CVPR-2019} and DeepFactors~\cite{Czarnowski-et-al-RAL-2020} learn a 2.5D depth representation and its probabilistic fusion rather than fusing into a full 3D model.
The DeepTAM~\cite{Zhou-et-al-IJCV-2020} mapping algorithm builds upon traditional cost volume computation with hand-crafted photoconsistency measures, which are fed into a neural network to estimate depth, but full 3D model fusion is not considered.
DeFuSR~\cite{Donne-Geiger-CVPR-2019} refines depth maps by improving cross-view consistency via reprojection, but it is not real-time capable.
Similar to our approach, Weder~\etal~\cite{Weder-et-al-CVPR-2020} perform online reconstruction and learn the fusion updates and weights. 
In contrast to our work, all information is fused into an SDF representation which requires handcrafted pre- or post-filtering to handle outliers, which is not end-to-end trainable.
The recent {ATLAS}~\cite{Murez-et-al-ECCV-2020} method fuses features from RGB input into a voxel grid and then regresses a TSDF volume.
While our method learns the fusion of features, they use simple weighted averaging. 
Their large ResNet50 backbone limits real-time capabilities.

\boldparagraph{Sequence to Vector Learning.}
On a high-level, our method processes a variable length sequence of depth maps and learns a 3D shape represented by a fixed length vector. 
It is thus loosely related to areas like video representation learning~\cite{Srivastava-et-al-ICML-2015} or sentiment analysis \cite{Medhat-et-al-AinShams-2015,Zhang-et-al-WIRE-2018} which processes text, audio or video to estimate a single vectorial value. 
In contrast to these works, we process 3D data and explicitly model spatial geometric relationships.%

\section{Method}

\begin{figure*}[t]
	\includegraphics[width=\linewidth]{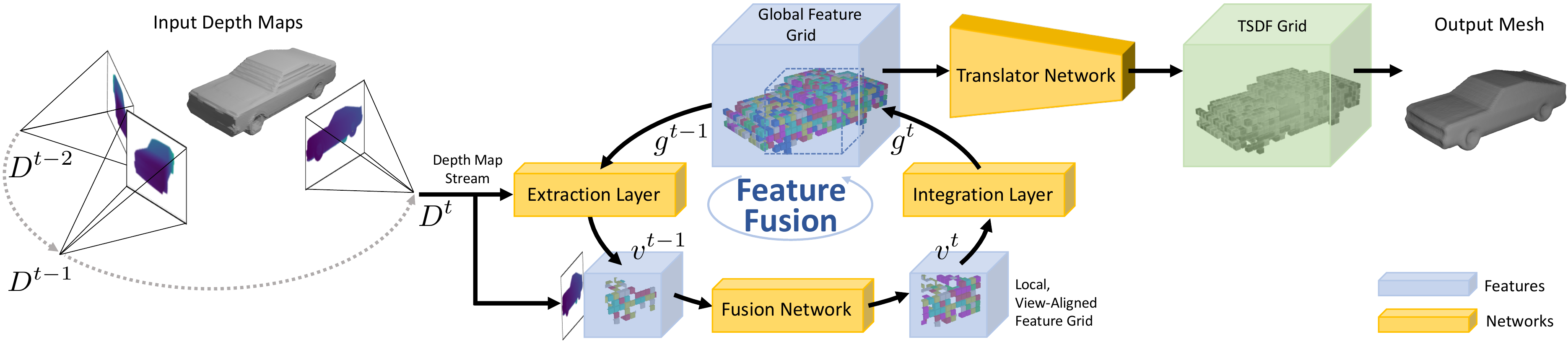}
	\caption{\textbf{Proposed online reconstruction approach.} Our pipeline consists of two main parts: \textbf{1)} A fusion network with its extraction and integration layers, and \textbf{2)} A translator network that translates the feature representation into an interpretable TSDF representation. 
		For any new depth map $\nDepth^t$ a local, view-aligned feature grid $\nLFeatGrid^{t-1}$ is extracted from the previous global feature grid $\nFeatGrid^{t-1}$.
		The fusion network updates the local feature grid $\nLFeatGrid^t$ which is then integrated back into an updated global feature grid $\nFeatGrid^t$.
		The translator network is independent of the fusion process and can be used asynchronously for an efficient fusion process.}
	\label{fig:pipeline_overview}
\end{figure*}
\boldparagraph{Overview.}
Given a stream of input depth maps $\nDepth^t: \nR^2 \rightarrow \nR$ with known camera calibration for each time step $t \in \nN$, we aim to fuse all surface information into a globally consistent scene representation  $\nFeatGrid: \nR^3 \rightarrow \nR^\nFeatDim$ while removing noise and outliers as well as complete potentially missing observations.
The final output of our method is a TSDF map $\nTSDF: \nR^3 \rightarrow \nR$, which can be processed into a mesh with standard iso-surface extraction methods as well as an occupancy map $\nOcc: \nR^3 \rightarrow [0,1]$. %
Figure~\ref{fig:pipeline_overview} provides an overview of our method.
The key idea is to decouple the scene representation for geometry fusion from the output scene representation. 
This decoupling is motivated by the difficulty for existing methods to handle outliers within an online fusion method. %
Therefore, we propose to fuse geometric information into a latent feature space without any preliminary outlier pre-filtering.
A subsequent translator network then decodes the latent feature space into the output scene representation (\eg, a TSDF grid). 
This approach allows for better and end-to-end trainable handling of outliers and avoids any handcrafted post-filtering, which is inherently difficult to tune and typically decreases the completeness of the reconstruction.
Furthermore, the learned latent representation also enables to capture complex and higher resolution shape information, leading to more accurate reconstruction results.

Our feature fusion pipeline consists of four key stages depicted as networks in Figure~\ref{fig:pipeline_overview}.
The first stage extracts the current state of the global feature volume into a local, view-aligned feature volume using an affine mapping defined by the given camera parameters.
After the extraction, this local feature volume is passed together with the new depth measurement and the ray directions through a feature fusion network.
This feature fusion network predicts optimal updates for the local feature volume, given the new measurement and its old state.
The updates are integrated back into the global feature volume using the inverse affine mapping defined in the first stage.
These three stages form the core of the fusion pipeline and are executed iteratively on the input depth map stream.
An additional fourth stage translates the feature volume into an application-specific scene representation, such as a TSDF volume, from which one can finally render a mesh for visualization.
We detail our pipeline in the following and we refer to the supplementary material for additional low-level architectural details.

\boldparagraph{Feature Extraction.}
The goal of iteratively fusing depth measurements is to
\textbf{(a)} fuse information about previously unknown geometry,
\textbf{(b)} increase the confidence about already fused geometry, and %
\textbf{(c)} to correct wrong or erroneous entries in the scene.
Towards these goals, the fusion process takes the new measurements to update the previous scene state $\nFeatGrid^{t-1}$, which encodes all previously seen geometry.
For a fast depth integration, we extract a local view-aligned feature subvolume $\nLFeatGrid^{t-1}$ with one ray per depth measurement centered at the measured depth via nearest neighbor search in the grid positions.
Each ray of features in the local feature volume is concatenated with the ray direction and the new depth measurement.
This feature volume is then passed to the fusion network.

\boldparagraph{Feature Fusion.}
The fusion network fuses the new depth measurements $\nDepth^t$ into the existing local feature representation $\nLFeatGrid^{t-1}$.
Therefore, we pass the feature volume through four convolutional blocks to encode neighborhood information from a larger receptive field.
Each of these encoding blocks, consists of two convolutional layers with a kernel size of three. 
These layers are followed by layer normalization and tanh activation function. 
We found layer normalization to be crucial for training convergence.
The output of each block is concatenated with its input, thereby generating a successively larger feature volume with increasing receptive field.
The decoder then takes the output of the four encoding blocks to predict feature updates.%
The decoder consists of four blocks with two convolutional layers and interleaved layer normalization and tanh activation.
The output of the final layer is passed through a single linear layer.
Finally, the predicted feature updates are normalized and passed as $\nLFeatGrid^t$ to the feature integration.

\boldparagraph{Feature Integration.}
The updated feature state is integrated back into the global feature grid by using the inverse global-local grid correspondences of the extraction mapping.
Similar to the extraction, we write the mapped features into the nearest neighbor grid location.
Since this mapping is not unique, we aggregate colliding updates using an average pooling operation.
Finally, the pooled features are combined with old ones using a per-voxel running average operation, where we use the update counts as weights.
This residual update operation ensures stable training and a homogeneous latent space, as compared to direct prediction of the global features.
Both the feature extraction and integration steps are inspired by \cite{Weder-et-al-CVPR-2020}, but they use tri-linear interpolation instead of nearest-neighbor sampling. 
When extracting and integrating features instead of SDF values, we empirically found that nearest-neighbor interpolation produces better results and leads to more stable convergence during training.

\boldparagraph{Feature Translation.}
\begin{figure*}
	\scriptsize
	\hspace{-1em}
	\begin{tabular}{cc@{\hspace{10pt}}c}
	  \includegraphics[width=0.40\linewidth]{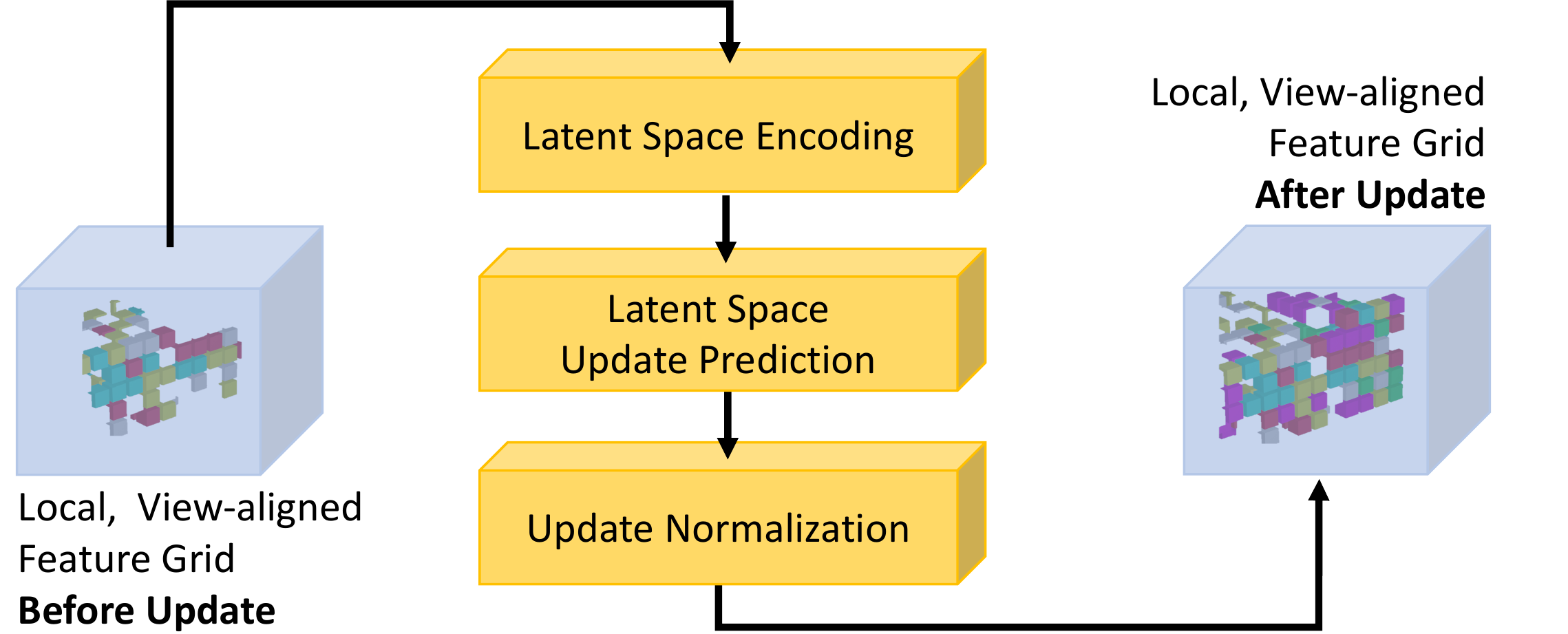} & &
	  \includegraphics[width=0.54\linewidth]{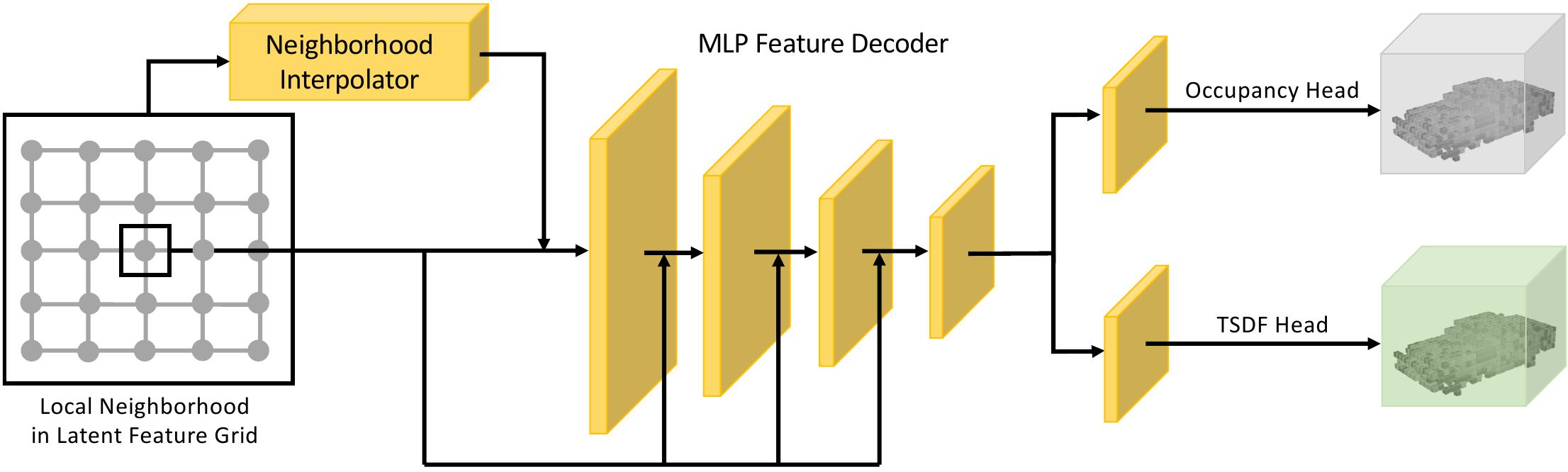} \\[1pt]
	  Fusion Network & & Translator Network \\[-6pt]
	\end{tabular}
	\caption{\textbf{Left:} The feature \textbf{fusion network} consists of a latent space encoder that fuses information from neighboring rays. This is followed by a latent updated predictor that predicts the updates for the latent space. Finally, the predicted features are normalized along the feature vector dimension.
	\textbf{Right:} The \textbf{translator network} consists of a series of neural blocks with linear layers, channel-wise dropout, and tanh activations. The first block extracts neighborhood information that is concatenated with the central feature vector. From the concatenated features the TSDF value is then predicted. All joining arrows correspond to a concatenation operation.}
\end{figure*}
In the final and possibly asynchronous step, we translate the latent scene representation $\nFeatGrid^t$ into a representation usable for visualization of the scene (\eg, signed distance field or occupancy grid). %
The network architecture in this step is inspired by IM-Net~\cite{Chen-et-al-CVPR-2019}.
For efficient and complete translation, we sample a regular grid of world coordinates.
Then, at each of these sampled points $\nQueryPoint_i$, the translator aggregates the information stored in the features of the local neighborhood and predicts the TSDF $\nTSDF(\nQueryPoint_i)$ as well as occupancy $\nOcc(\nQueryPoint_i)$ for this specific grid location.
To this end, the translator concatenates the feature vectors of the $5\times5\times5$ neighborhood and compresses them into a single feature vector using a linear layer followed by tanh activation.
Next, the so combined features are concatenated with the query point feature $\nFeatGrid^t(\nQueryPoint_i)$ and passed through the remaining translation network, which consists of four linear layers interleaved with tanh activations and channel-wise dropout preventing the network from overfitting to a single feature channel.
According to the desired output ranges, the TSDF head is activated using tanh, while the occupancy head uses a sigmoid activation.
After each layer, we concatenate the output with the original query point feature $\nFeatGrid^t(\nQueryPoint_i)$.

\boldparagraph{Training Procedure and Loss Function.}
All networks are jointly trained end-to-end.
In each training epoch, we randomly shuffle the input depth maps and iteratively fuse them one by one into the corresponding latent feature grid $\nFeatGrid: \nR^3 \rightarrow \nR^\nFeatDim$.
After integrating the depth map into the latent feature grid, we query the translator network, where the latent feature grid was just updated, and render the TSDF $\nTSDF: \nR^3 \rightarrow\nR$ and occupancy $\nOcc: \nR^3 \rightarrow [0,1]$.
The entire pipeline is optimized using the following loss function: %

\begin{align}
\nonumber\\[-37pt]
\label{eq:loss-total}
\nLoss = &\frac{1}{n} \sum_{i}
	\lambda_1 \nLoss_1(\nTSDF_i, \hat{\nTSDF}_i) + 
	\lambda_2 \nLoss_2(\nTSDF_i, \hat{\nTSDF}_i) \nonumber\\ & +
	\lambda_{\nOcc} \nLoss_{\nOcc}(\nOcc_i, \hat{\nOcc}_i) + \lambda_{\nFeatGrid}\overline{\sigma_{ch}^2(\nFeatGrid)} \enspace ,
\end{align}
where $\mathcal{L}_1$ and $\mathcal{L}_2$ denote the $L_1$ and $L_2$ norms, and $\mathcal{L}_o$ is the binary cross-entropy on the predicted occupancy.
The $\mathcal{L}_2$ loss is helpful with outliers, whereas the $\mathcal{L}_1$ loss improves the reconstruction of fine details.
In each step, $n$ denotes the number of all updated feature grid locations. 
When training with outlier contaminated data, we found that setting $n$ equal to all visited feature grid locations yields the best results.
Therefore, $n$ is a crucial hyperparameter when training the pipeline.
Moreover, $\hat{\nTSDF}_i$ and $\hat{\nOcc}_i$ denote the ground-truth TSDF and occupancy value, respectively.
To avoid large deviations for a single feature in the latent space, we regularize the feature grid $\nFeatGrid$ by penalizing the mean of the channel-wise variance by $\overline{\sigma_{ch}^2(\nFeatGrid)}$.
We empirically set the loss weights to $\lambda_1=1.$, $\lambda_2=10.$, $\lambda_{\nOcc}=0.01$, and $\lambda_{\nFeatGrid}=0.05$.

\section{Experiments}
We first discuss implementation details and evaluation metrics before evaluating our method on synthetic and real-world data in comparison to other methods.
We further analyze our method for varying numbers of features $\nFeatDim$ in an ablation study.
We provide additional experiments and results in the supplementary material.

\boldparagraph{Implementation Details.}
Our pipeline is implemented in PyTorch and trained on an NVIDIA RTX 2080.
All networks were trained using the Adam optimizer~\cite{Kingma-Ba-ICLR-2014} with an initial learning rate of $0.01$, which was adapted using an exponential learning rate scheduler at a rate of $0.998$.
For momentum and beta, we empirically found the default parameters to yield the best results.
We trained all networks on synthetic data being augmented with artificial noise and outliers.
The batch-size is set to one due to the nature of the sequential fusion process.
However, we accumulate the gradients across 8 scene update steps and then update the network parameters.
Our un-optimized implementation runs at $\sim\!7$ frames per seconds with a depth map resolution of $240\times320$ on an NVIDIA RTX 2080.
This demonstrates the real-time applicability of our approach. %

\boldparagraph{Evaluation Metrics.}
We use the following evaluation metrics to quantify the performance of our approach: Mean Squared Error (\textbf{MSE}), Mean Absolute Distance (\textbf{MAD}), Accuracy (\textbf{Acc.}), Intersection-over-Union (\textbf{IoU}), Mesh Completeness (\textbf{M.C.}) Mesh Accuracy (\textbf{M.A.}), and \textbf{F1} score. 
Further details are in the supplementary material.

\subsection{Results on Synthetic Data}

\boldparagraph{Datasets.}
We used the synthetic ShapeNet~\cite{Chang-et-al-Arxiv-2015} and ModelNet~\cite{Wu-et-al-CVPR-2015} datasets for performance evaluation.
From ShapeNet, we selected 13 classes for training and evaluate on the same test set as~\cite{Weder-et-al-CVPR-2020} consisting of 60 objects from six classes, for which pretrained models~\cite{Park-et-al-CVPR-2019} are available.
For ModelNet~\cite{Wu-et-al-CVPR-2015}, we trained and tested on 10 classes using the train-test split from~\cite{Weder-et-al-CVPR-2020}.
We first generated watertight models using the mesh-fusion pipeline used in~\cite{Mescheder-et-al-CVPR-2019} and computed TSDFs using the mesh-to-sdf\footnote{\url{https://github.com/marian42/mesh_to_sdf}} library.
Additionally, we render depth frames for 100 randomly sampled camera views for each mesh.
These depth maps are the input to our pipeline and existing methods.
For both datasets, we found that training on one single object per class is sufficient for generalization to any other object and class.

\begin{figure}[t]
	\centering
	\scriptsize
	\begingroup	 %
	\renewcommand{\arraystretch}{1.1}
	\setlength{\tabcolsep}{7.4pt}
	\begin{tabular}{lrrrrr}
		\toprule
		\textbf{Method} & \textbf{MSE}$\downarrow$ & \textbf{MAD}$\downarrow$ & \textbf{Acc.}$\uparrow$ & \textbf{IoU}$\uparrow$ & \textbf{F1}$\uparrow$ \\
		& {[}e-05{]} & {[}e-02{]} & [\%] & [0,1] & [0,1] \\
		\midrule
		DeepSDF~\cite{Park-et-al-CVPR-2019}            & 464.0 & 4.99 & 66.48 & 0.538 & 0.66 \\
		Occ.Net.~\cite{Mescheder-et-al-CVPR-2019}      & 56.8  & 1.66 & 85.66 & 0.484 & 0.62 \\
		IF-Net~\cite{Chibane-et-al-CVPR-2020}          & 6.2   & 0.47 & 93.16 & 0.759 & 0.86 \\
		\midrule
		TSDF Fusion~\cite{Curless-et-al-SIGGRAPH-1996} & 11.0  & 0.78 & 88.06 & 0.659 & 0.79 \\
		TSDF + 2D denoising                            & 27.0  & 0.84 & 87.48 & 0.650 & 0.78 \\
		TSDF + 3D denoising                            & 8.2   & 0.61 & 94.76 & 0.816 & 0.89 \\
		RoutedFusion~\cite{Weder-et-al-CVPR-2020}      & 5.9   & 0.50 & 94.77 & 0.785 & 0.87 \\
		Ours & \textbf{2.9} & \textbf{0.27} & \textbf{97.00} & \textbf{0.890} & \textbf{0.94} \\
		\bottomrule
	\end{tabular}
	\endgroup
	\begingroup
	\setlength{\tabcolsep}{2pt}
	\newcommand{\sz}{0.264}
	\newcommand{\szz}{0.143}
	\hspace{-9pt}
	\begin{tabular}{cccccc}
		\\[-2pt]
		\includegraphics[height=\sz\columnwidth]{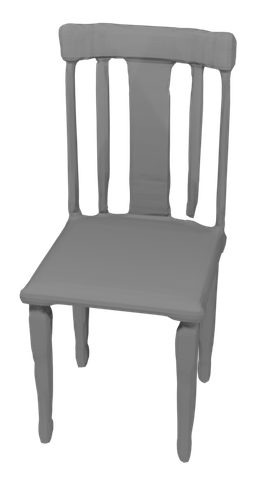} &
		\includegraphics[height=\sz\columnwidth]{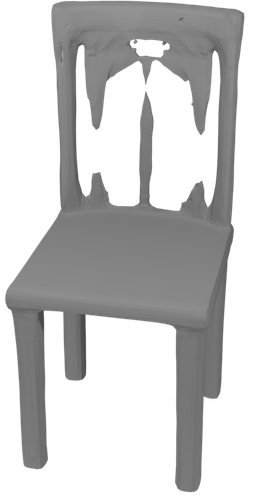} &
		\includegraphics[height=\sz\columnwidth]{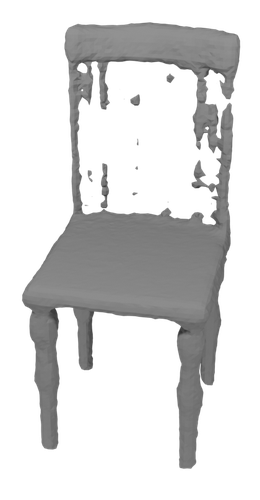} &
		\includegraphics[height=\sz\columnwidth]{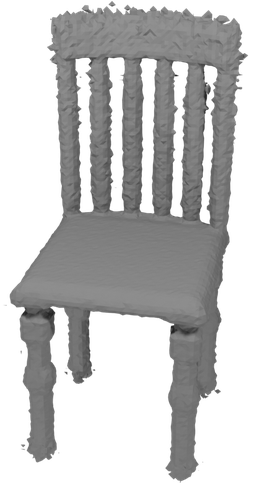} &
		\includegraphics[height=\sz\columnwidth]{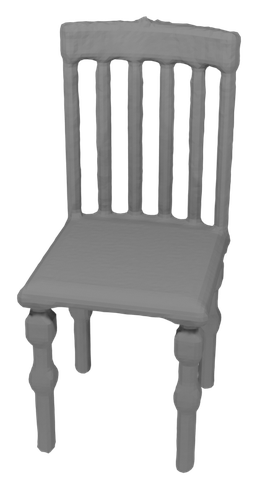} &
		\includegraphics[height=\sz\columnwidth]{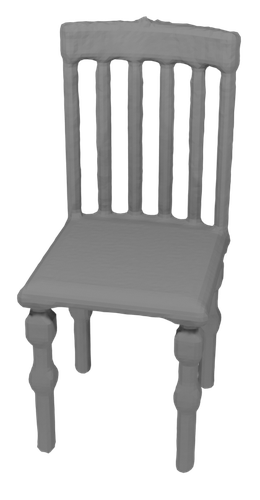} \\
		\includegraphics[width=\szz\columnwidth]{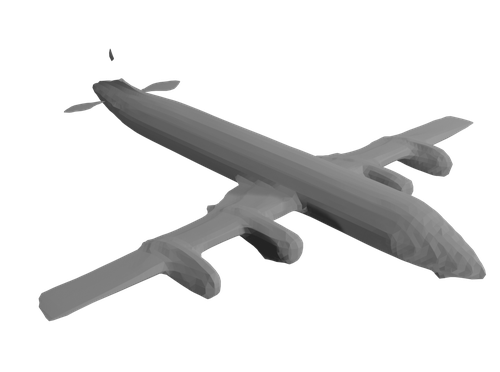} &
		\includegraphics[width=\szz\columnwidth]{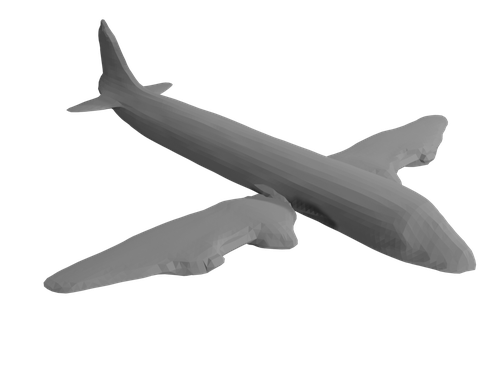} &
		\includegraphics[width=\szz\columnwidth]{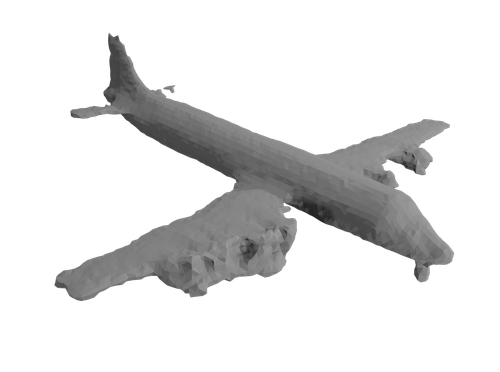} &
		\includegraphics[width=\szz\columnwidth]{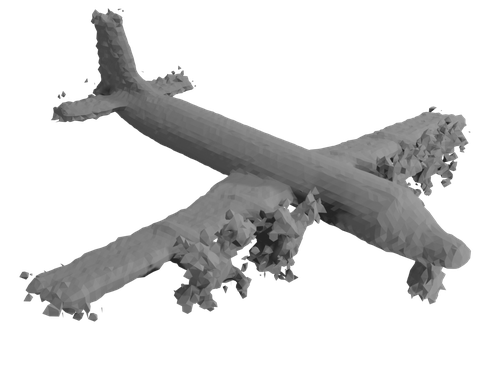} &
		\includegraphics[width=\szz\columnwidth]{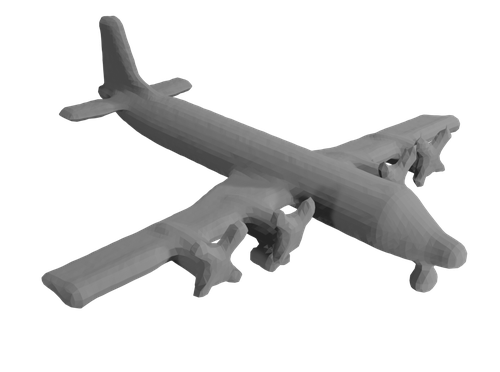} &
		\includegraphics[width=\szz\columnwidth]{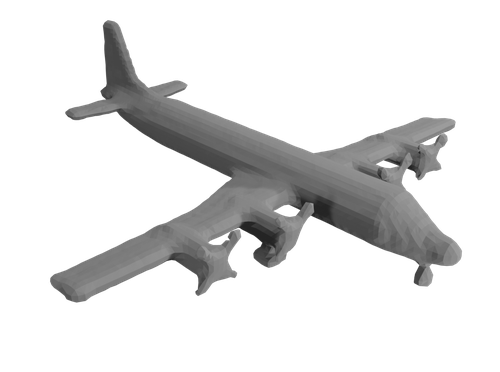} \\
		DeepSDF~\cite{Park-et-al-CVPR-2019} & Occ.Net.~\cite{Mescheder-et-al-CVPR-2019} & IF-Net~\cite{Chibane-et-al-CVPR-2020} & TSDF  & Routed- & Ours \\
		& & & Fusion~\cite{Curless-et-al-SIGGRAPH-1996} & Fusion~\cite{Weder-et-al-CVPR-2020} & \\[-5pt] 
	\end{tabular}
	\endgroup
	\caption{\textbf{Quantitative and qualitative results on ShapeNet~\cite{Chang-et-al-Arxiv-2015}.} Our fusion approach consistently outperforms all baselines and state of the art in both, scene representation and depth map fusion. The performance differences to~\cite{Weder-et-al-CVPR-2020} are also visualized in~\figurename{}~\ref{fig:error-maps}.}
	\label{fig:existing-method-comparison}	
\end{figure}

\begin{figure}[t]	
	\centering
	\tiny
	\setlength{\tabcolsep}{3pt}
	\renewcommand{\arraystretch}{1.1}
	\newcommand{\sz}{0.138}
	\begin{tabular}{ccccccc}
		\raisebox{0.01\height }{\rotatebox{90}{TSDF Fusion}} &
		\includegraphics[height=\sz\columnwidth]{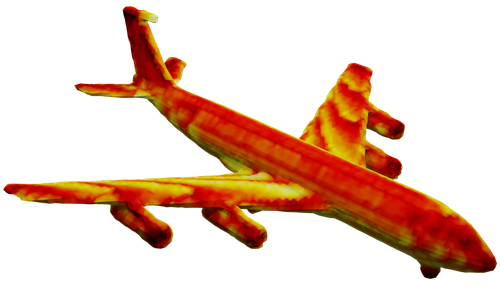} &
		\includegraphics[height=\sz\columnwidth]{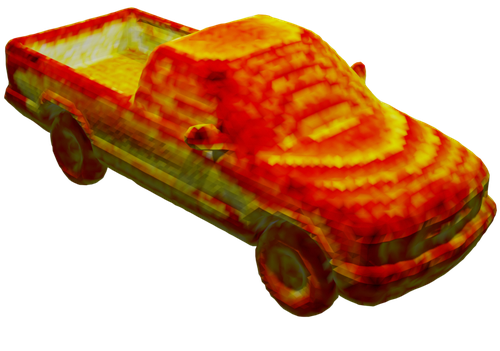} &
		\includegraphics[height=\sz\columnwidth]{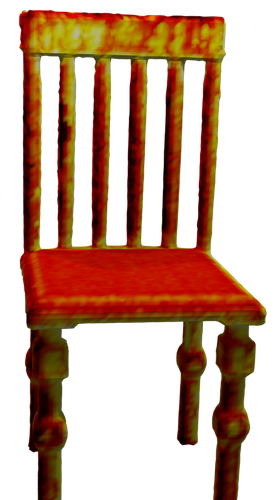} &
		\includegraphics[height=\sz\columnwidth]{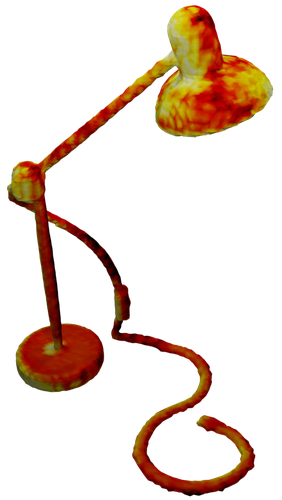} &
		\includegraphics[height=\sz\columnwidth]{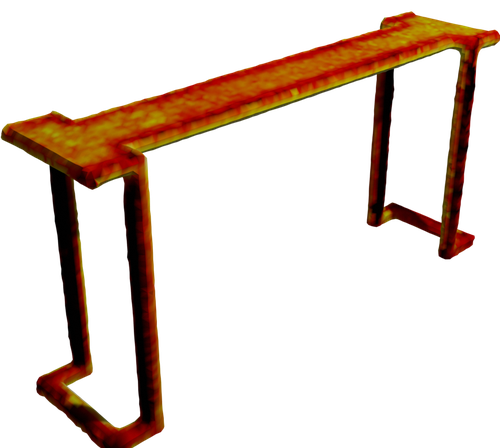} &
		\multirow[c]{-3}{*}[0.22cm]{\includegraphics[height=3cm]{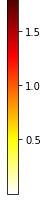}} \\
		\raisebox{0.01\height }{\rotatebox{90}{RoutedFusion}} &
		\includegraphics[height=\sz\columnwidth]{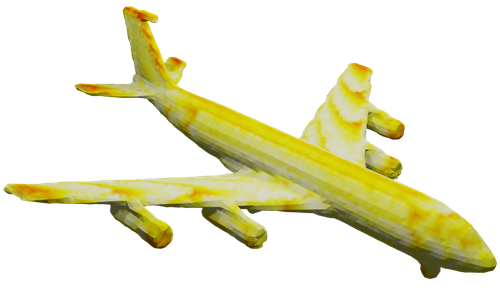} &
		\includegraphics[height=\sz\columnwidth]{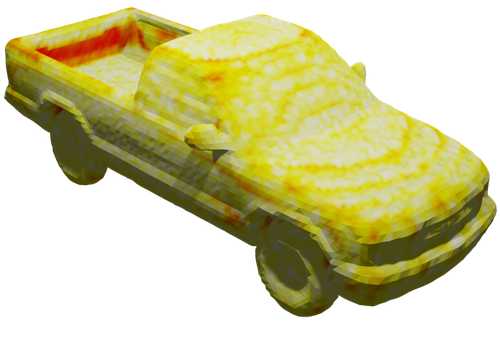} &
		\includegraphics[height=\sz\columnwidth]{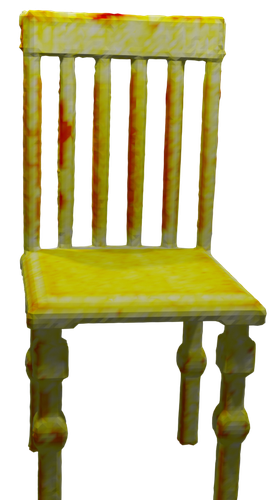} &
		\includegraphics[height=\sz\columnwidth]{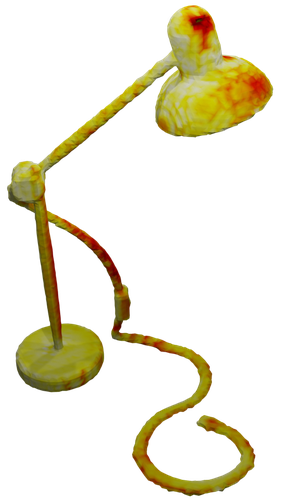} &
		\includegraphics[height=\sz\columnwidth]{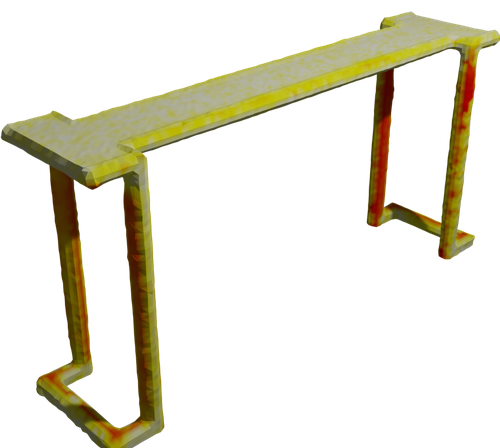} \\
		\raisebox{1.1\height }{\rotatebox{90}{Ours}} &
		\includegraphics[height=\sz\columnwidth]{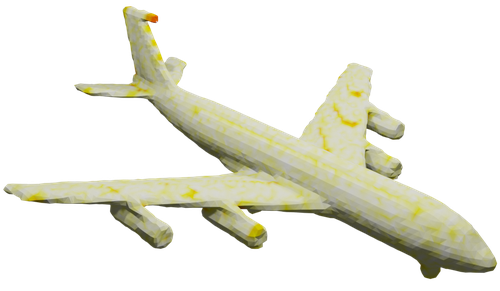} &
		\includegraphics[height=\sz\columnwidth]{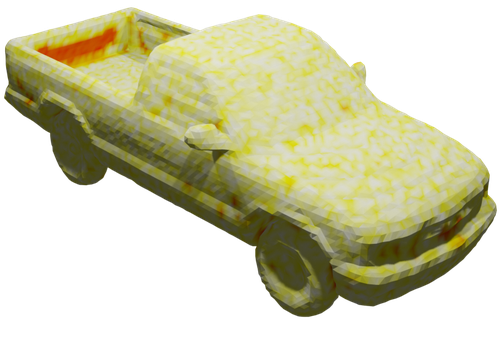} &
		\includegraphics[height=\sz\columnwidth]{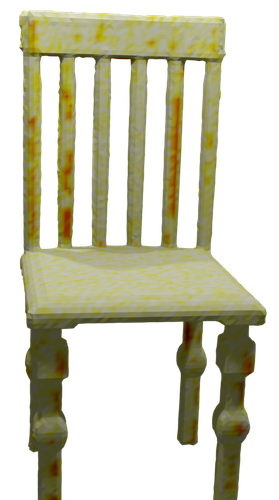} &
		\includegraphics[height=\sz\columnwidth]{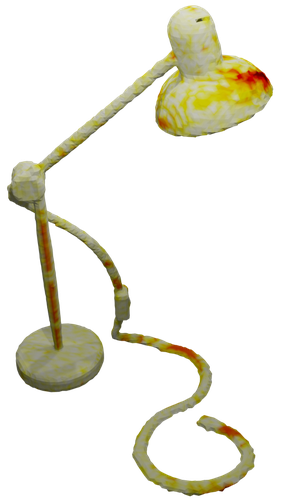} &
		\includegraphics[height=\sz\columnwidth1]{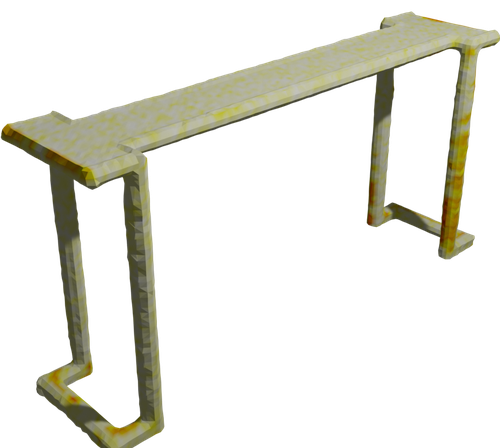} \\[-5pt] 
	\end{tabular}
	\caption{\textbf{Mesh Accuracy (M.A.) visualization on ShapeNet meshes.} Our method consistently reconstructs more accurate meshes than the baseline depth fusion methods. Especially thin geometries (table/chair legs, lamp cable) are reconstructed with better accuracy.}
	\label{fig:error-maps}	
\end{figure}

\boldparagraph{Comparison to Existing Methods.}
For performance comparisons, we fuse depth maps and augment them with artificial depth-dependent noise as in~\cite{Riegler-et-al-3DV-2017}.
We compare to state-of-the-art learned scene representation methods DeepSDF~\cite{Park-et-al-CVPR-2019},  OccupancyNetworks~\cite{Mescheder-et-al-CVPR-2019}, and IF-Net~\cite{Chibane-et-al-CVPR-2020}, as well as to the online fusion methods TSDF Fusion~\cite{Curless-et-al-SIGGRAPH-1996} and RoutedFusion~\cite{Weder-et-al-CVPR-2020}.
We further implemented two additional baselines to demonstrate the benefits of a fully learned scene representation for depth map fusion: 
\textbf{(1)} one baseline performs a learned 2D noise filtering before fusing the frames using TSDF Fusion~\cite{Curless-et-al-SIGGRAPH-1996}, and 
\textbf{(2)} a baseline that post-processes models fused by TSDF Fusion using a simplification of our translation network - the principle is similar to OctnetFusion~\cite{Riegler-et-al-3DV-2017}, but on a dense grid. 
Further details on these baselines is given in the supplementary material.
We compare all baselines on the test set of Weder~\etal~\cite{Weder-et-al-CVPR-2020} in \figurename~\ref{fig:existing-method-comparison}. %
For input data augmentation, we used the same scale $0.005$ as in~\cite{Weder-et-al-CVPR-2020}.
\figurename~\ref{fig:existing-method-comparison} shows that our method significantly outperforms all existing depth map fusion as well as learned scene representations.
We especially emphasize the increase in IoU by more than 10\%.
This significant increase is due to many fine-grained improvements, where RoutedFusion~\cite{Weder-et-al-CVPR-2020} wrongly predicts the sign, as shown in \figurename~\ref{fig:error-maps}.
In all experiments, we set the truncation distance of TSDF Fusion to $4$cm, which is similar to the receptive field of our fusion network.

\begin{figure}
	\centering
	\hspace{-10pt}
	\begin{tabular}{cc}
		\includegraphics[width=0.48\columnwidth]{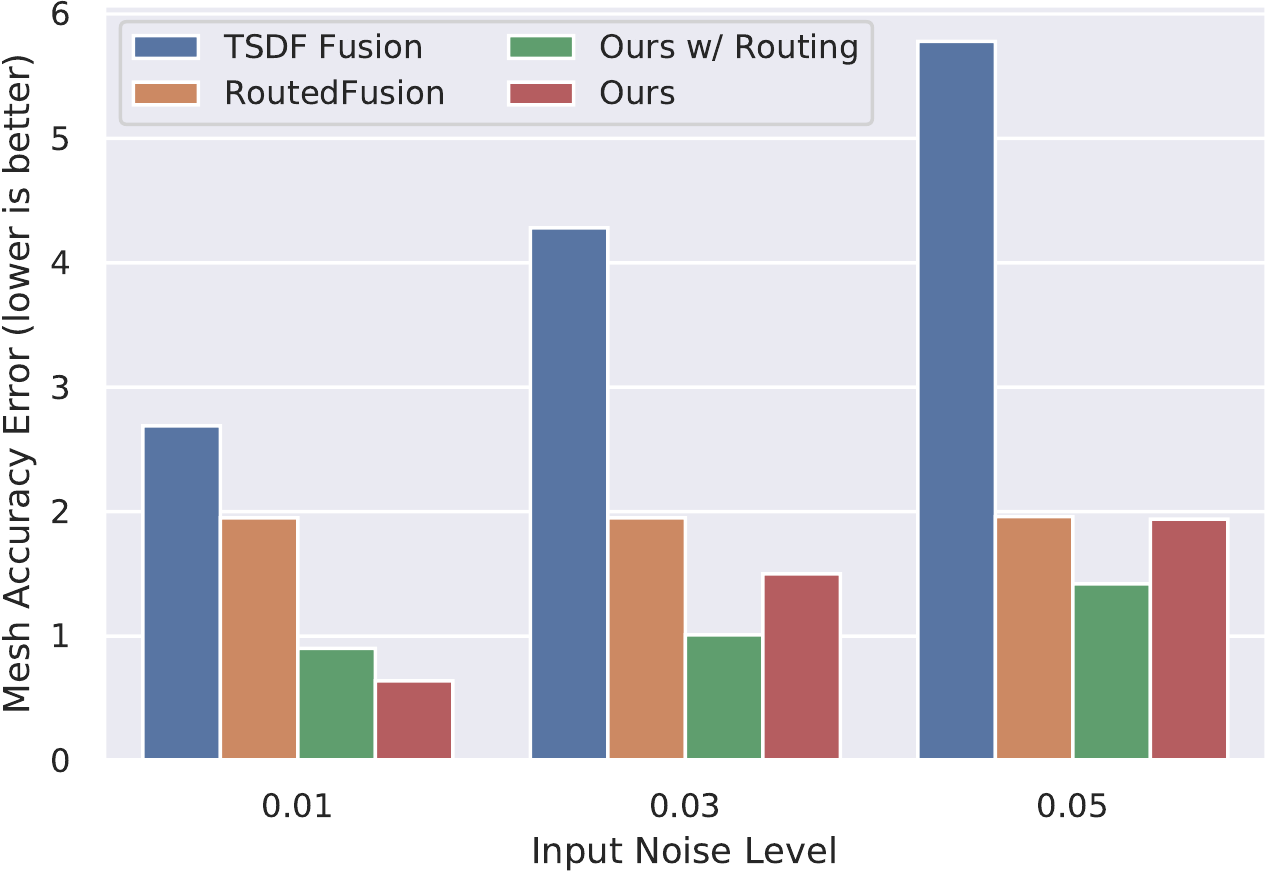} & \hspace{-13pt}
		\includegraphics[width=0.49\columnwidth]{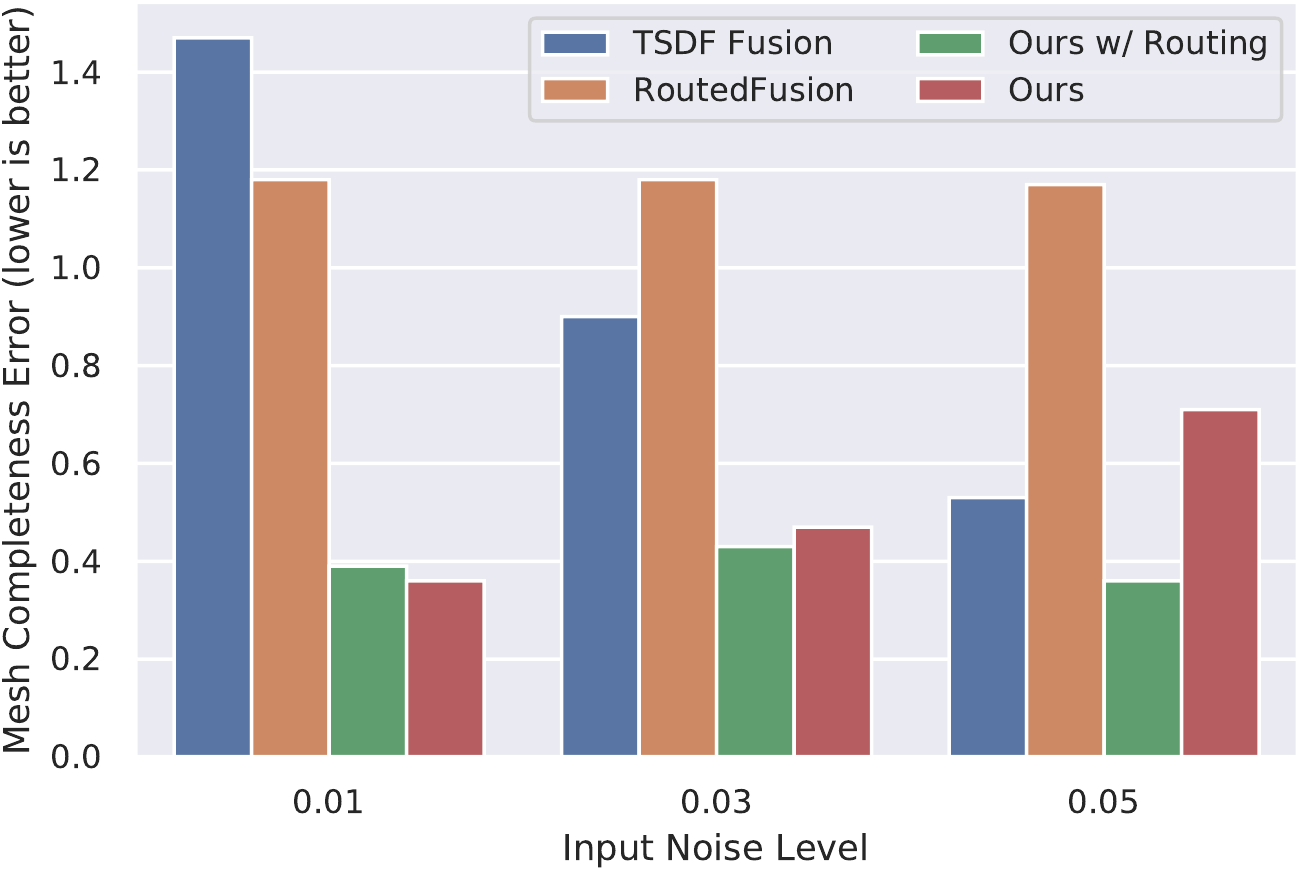}\\[-0.2cm]
	\end{tabular}
	\caption{\textbf{Reconstruction from Noisy Depth Maps.} Our method outperforms existing depth fusion methods for various input noise levels. The performance can be further boosted by preprocessing the depth maps with the routing network from~\cite{Weder-et-al-CVPR-2020} leading to better robustness to high input noise levels, while it over-smoothes in the absence of noise.}
	\label{fig:noise-experiment}
\end{figure}

\boldparagraph{Higher Input Noise Levels.}
We also assess our method in fusing depth maps corrupted with higher noise levels on the ModelNet dataset~\cite{Wu-et-al-CVPR-2015} in \figurename~\ref{fig:noise-experiment}.
For this experiment, we augment the input depth maps with three different noise levels.
We fuse the corrupted depth maps using standard TSDF Fusion~\cite{Curless-et-al-SIGGRAPH-1996} and RoutedFusion~\cite{Weder-et-al-CVPR-2020}.
Since Weder~\etal~\cite{Weder-et-al-CVPR-2020} showed that their proposed routing network significantly improved the robustness to higher input noise levels, we also tested our method with depth maps pre-processed by a routing network.
For these experiments, we use the pre-trained routing network provided by~\cite{Weder-et-al-CVPR-2020}.

\begin{figure*}[tbh]
  \hspace{-2.5em}  %
  \begin{minipage}[c]{0.4\textwidth}
	\centering
	\scriptsize
	\setlength{\tabcolsep}{5.3pt}
	\newcommand{\lsp}{8pt}	%
	\newcommand{\bsp}{2pt}	%
	\begin{tabular}{llrrrr}
		\toprule
		\textbf{Outlier} & \textbf{Method} & \textbf{MSE}$\downarrow$ & \textbf{MAD}$\downarrow$ & 
		\textbf{Acc.$\uparrow$} & \textbf{IoU}$\uparrow$ \\ %
		\textbf{Fraction} &    & [e-05] & [e-02] & [\%] & [0,1] \\
		\midrule \noalign{\vskip 3pt}
		\multirow{4}{*}{0.01}
		& \begin{tabular}{l}TSDF\\Fusion\end{tabular}    & 34.51 & 1.17  & 85.17 & 0.645 \\[\lsp]
		& \begin{tabular}{l}Routed-\\Fusion\end{tabular} & 5.43  & 0.57  & 95.21 & 0.837 \\[\lsp]
		& \begin{tabular}{l}Ours\end{tabular}            & \textbf{2.27} & \textbf{0.29} &  \textbf{97.57} & \textbf{0.884}\\[\bsp]
		\midrule \noalign{\vskip \bsp}
		\multirow{4}{*}{0.05}
		& \begin{tabular}{l}TSDF\\Fusion\end{tabular}    & 80.72 & 2.02 & 73.86 & 0.432 \\[\lsp]
		& \begin{tabular}{l}Routed-\\Fusion\end{tabular} & 9.84  & 0.68 & 94.46 & 0.803 \\[\lsp]
		& \begin{tabular}{l}Ours\end{tabular}            & \textbf{4.91} & \textbf{0.22} & \textbf{98.05} & \textbf{0.851} \\[\bsp]
		\midrule \noalign{\vskip \bsp}
		\multirow{4}{*}{0.1}
		& \begin{tabular}{l}TSDF\\Fusion\end{tabular}    &102.50 & 2.43 & 67.47 & 0.341 \\[\lsp]
		& \begin{tabular}{l}Routed-\\Fusion\end{tabular} & 14.25 & 0.77 & 92.95 & 0.764 \\[\lsp]
		& \begin{tabular}{l}Ours\end{tabular}            & \textbf{3.35} & \textbf{0.22} & \textbf{98.48} & \textbf{0.865} \\[2pt]
		\bottomrule
	\end{tabular}
  \end{minipage}
  \begin{minipage}[c]{0.55\textwidth}
	\centering
	\scriptsize
	\setlength{\tabcolsep}{0pt}
	\renewcommand{\arraystretch}{0.3}
	\newcommand{\sz}{0.21}
	\begin{tabular}{cccccc}
		\multicolumn{3}{c}{\textbf{Reconstructed Geometry}} & \multicolumn{3}{c}{\textbf{Outlier Projection onto XY-Plane}} \\[3pt]
		TSDF   & Routed-  &      & TSDF   & Routed-  &  \\[1pt]
		Fusion~\cite{Curless-et-al-SIGGRAPH-1996} & Fusion~\cite{Weder-et-al-CVPR-2020} & Ours &
		Fusion~\cite{Curless-et-al-SIGGRAPH-1996} & Fusion~\cite{Weder-et-al-CVPR-2020} & Ours\\[1pt]
		\includegraphics[height=\sz\columnwidth]{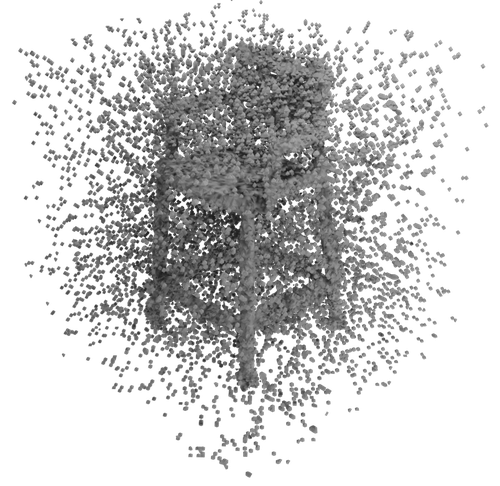} &
		\includegraphics[height=\sz\columnwidth]{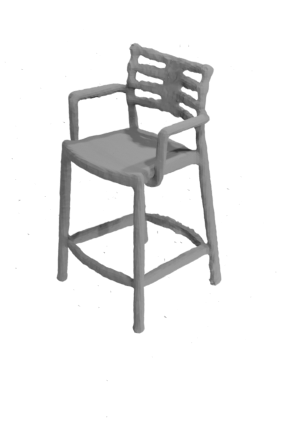} &
		\includegraphics[height=\sz\columnwidth]{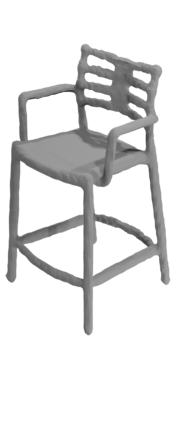} &
		\includegraphics[height=\sz\columnwidth]{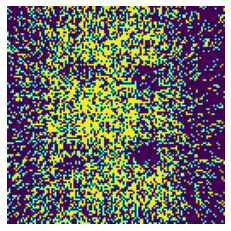} &
		\includegraphics[height=\sz\columnwidth]{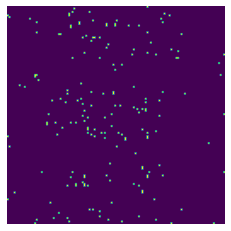} &
		\includegraphics[height=\sz\columnwidth]{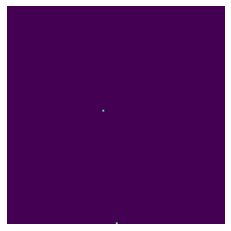} \\
		\includegraphics[height=\sz\columnwidth]{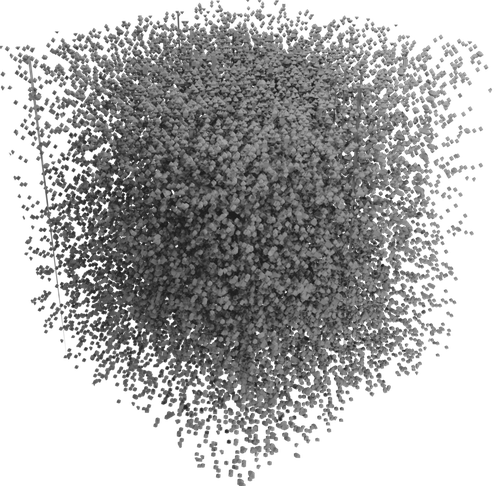} &
		\includegraphics[height=\sz\columnwidth]{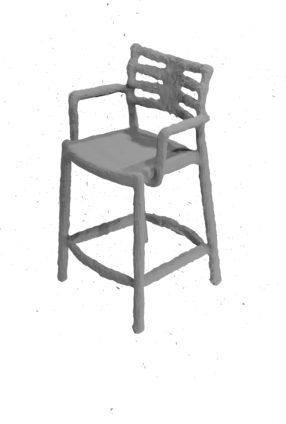} &
		\includegraphics[height=\sz\columnwidth]{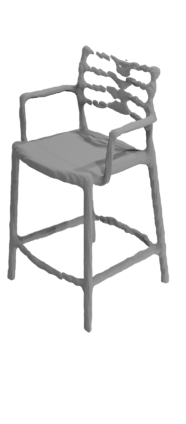} &
		\includegraphics[height=\sz\columnwidth]{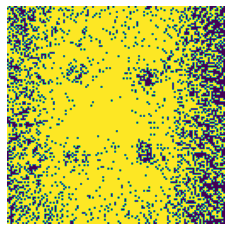} &
		\includegraphics[height=\sz\columnwidth]{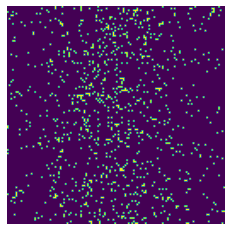} &
		\includegraphics[height=\sz\columnwidth]{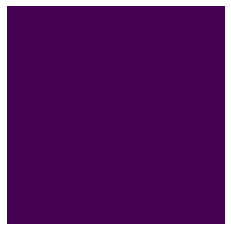} \\
		\includegraphics[height=\sz\columnwidth]{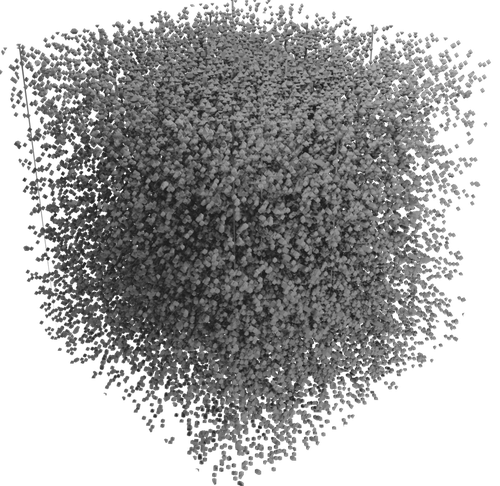} &
		\includegraphics[height=\sz\columnwidth]{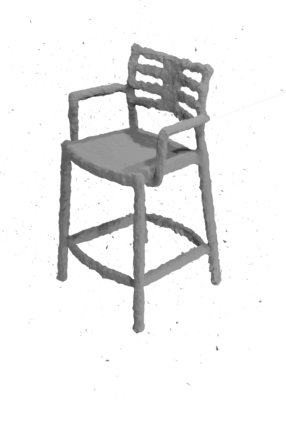} &
		\includegraphics[height=\sz\columnwidth]{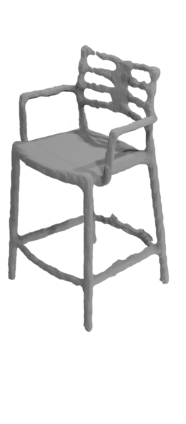} &
		\includegraphics[height=\sz\columnwidth]{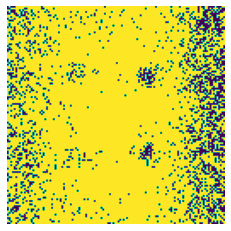} &
		\includegraphics[height=\sz\columnwidth]{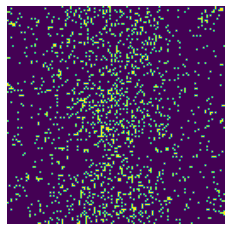} &
		\includegraphics[height=\sz\columnwidth]{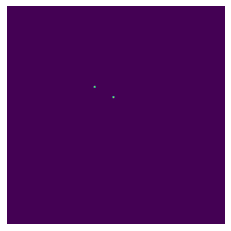} \\[3pt]
	\end{tabular}
  \end{minipage}
  \vspace{-6pt}
  \caption{\textbf{Reconstruction from Outlier-Contaminated Data.} The left table states performance measures for various outlier fractions. The figures on the right show corresponding reconstruction results and errors projected on the xy-plane for an exemplary ModelNet~\cite{Wu-et-al-CVPR-2015} model.
  Our method outperforms state-of-the-art depth fusion methods regardless of the outlier fraction, but in particular with larger outlier amounts. Note that high outlier rates are common in multi-view stereo as shown in the supp. material of~\cite{Knapitsch-et-al-TOG-2017}.}
  \label{fig:outlier-comparison}
\end{figure*}

\boldparagraph{Outlier Handling.}
A main drawback of~\cite{Weder-et-al-CVPR-2020} is its limitation in handling outliers.
To this end, we run an experiment, where we augment the input depth maps with random outlier blobs.
We create this data by sampling an outlier map from a fixed distribution scaled by a fixed outlier scale.
Additionally, we sample three masks with a given probability (outlier fraction) and dilate it once, twice, and three times, respectively.
Then, these masks are used to select the outliers from the outlier map.
We report the results of this experiment in \figurename~\ref{fig:outlier-comparison}.
Note that the results might be better with even higher outlier fractions since we only evaluate on updated grid locations. 
The consistency in outlier filtering and increase in updated grid locations improves the metrics.

\subsection{Ablation Study}
In a series of ablation studies, we discuss several benefits of our pipeline and justify design choices.

\begin{figure}[h!]
	\centering
	\newcommand{\sz}{0.49}
	\begin{tabular}{cc}
		\hspace{-6pt}\includegraphics[width=\sz\columnwidth]{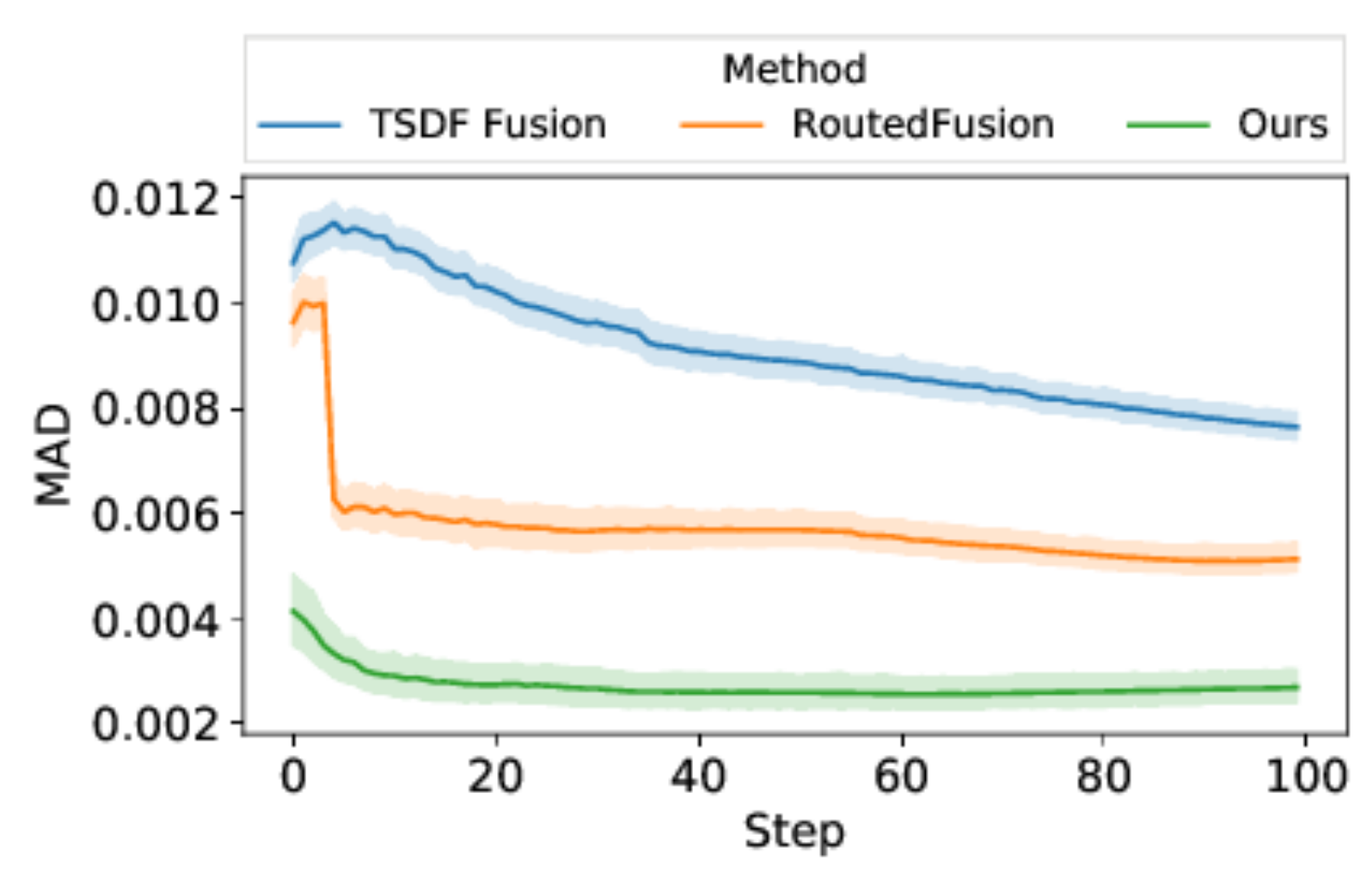} &
		\hspace{-8pt}\includegraphics[width=\sz\columnwidth]{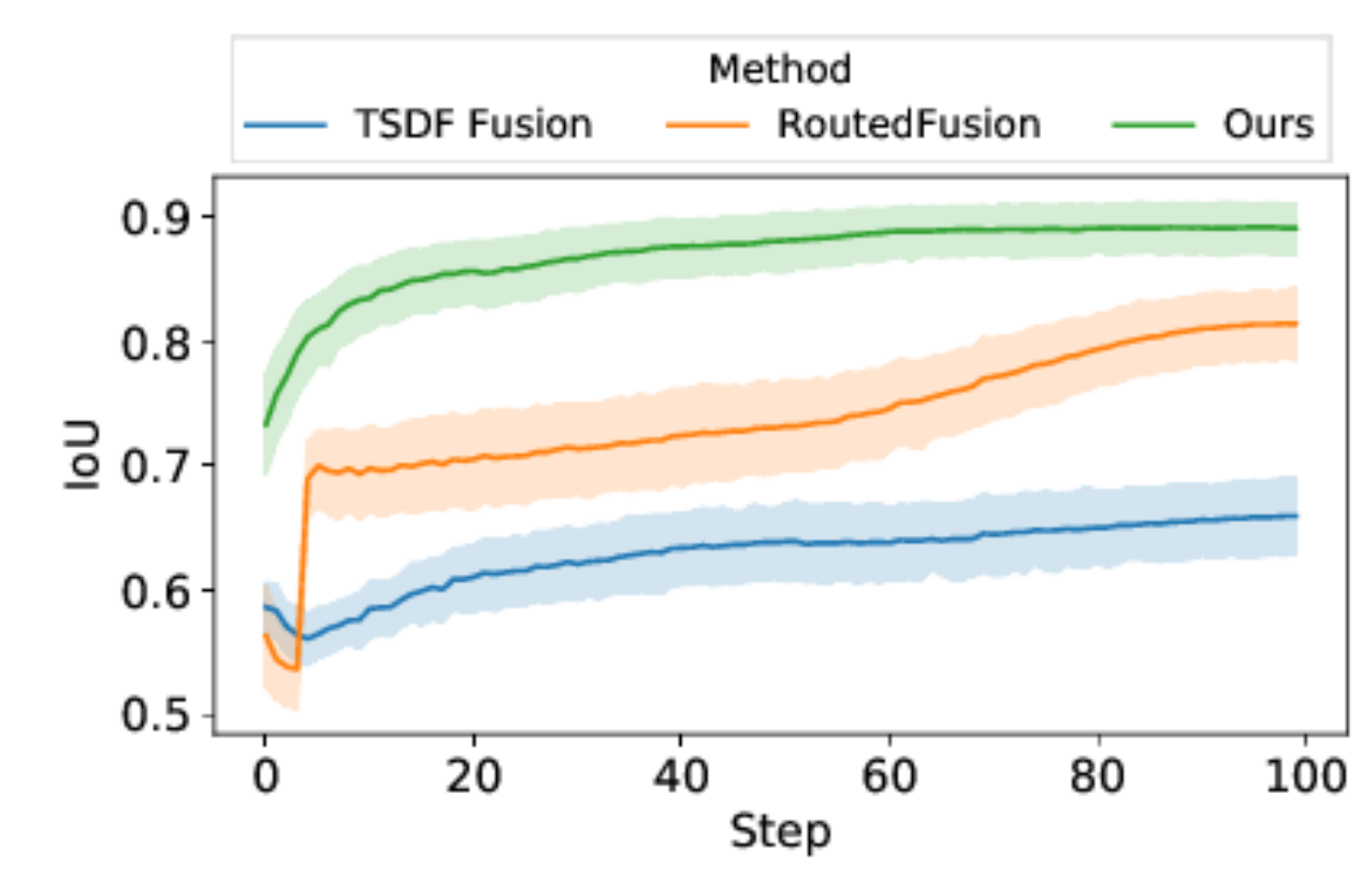} \\[-10pt]
	\end{tabular}
	\vspace{8pt}
	\caption{\textbf{Performance of iterative fusion over time.} Our method consistently outperforms both baselines~\cite{Weder-et-al-CVPR-2020, Curless-et-al-SIGGRAPH-1996} at every step of the fusion procedure.}
	\label{fig:iterative-fusion}
	\vspace{-10pt}
\end{figure}

\boldparagraph{Iterative Fusion.}
Ideally, fusion algorithms should be independent from the number of integrated frames and steadily improve the reconstruction as new information becomes available.
In \figurename{}~\ref{fig:iterative-fusion}, we show that our method is not only better than competing algorithms from the start, but also continuously improves the reconstruction as more data is fused.
The metrics are averaged at every fusion step over all scenes in the test set used for all experiments on ShapeNet~\cite{Chang-et-al-Arxiv-2015}.

\begin{figure}[h!]
	\centering
	\newcommand{\sz}{0.49}
	\begin{tabular}{cc}
		\hspace{-7pt}\includegraphics[width=\sz\columnwidth]{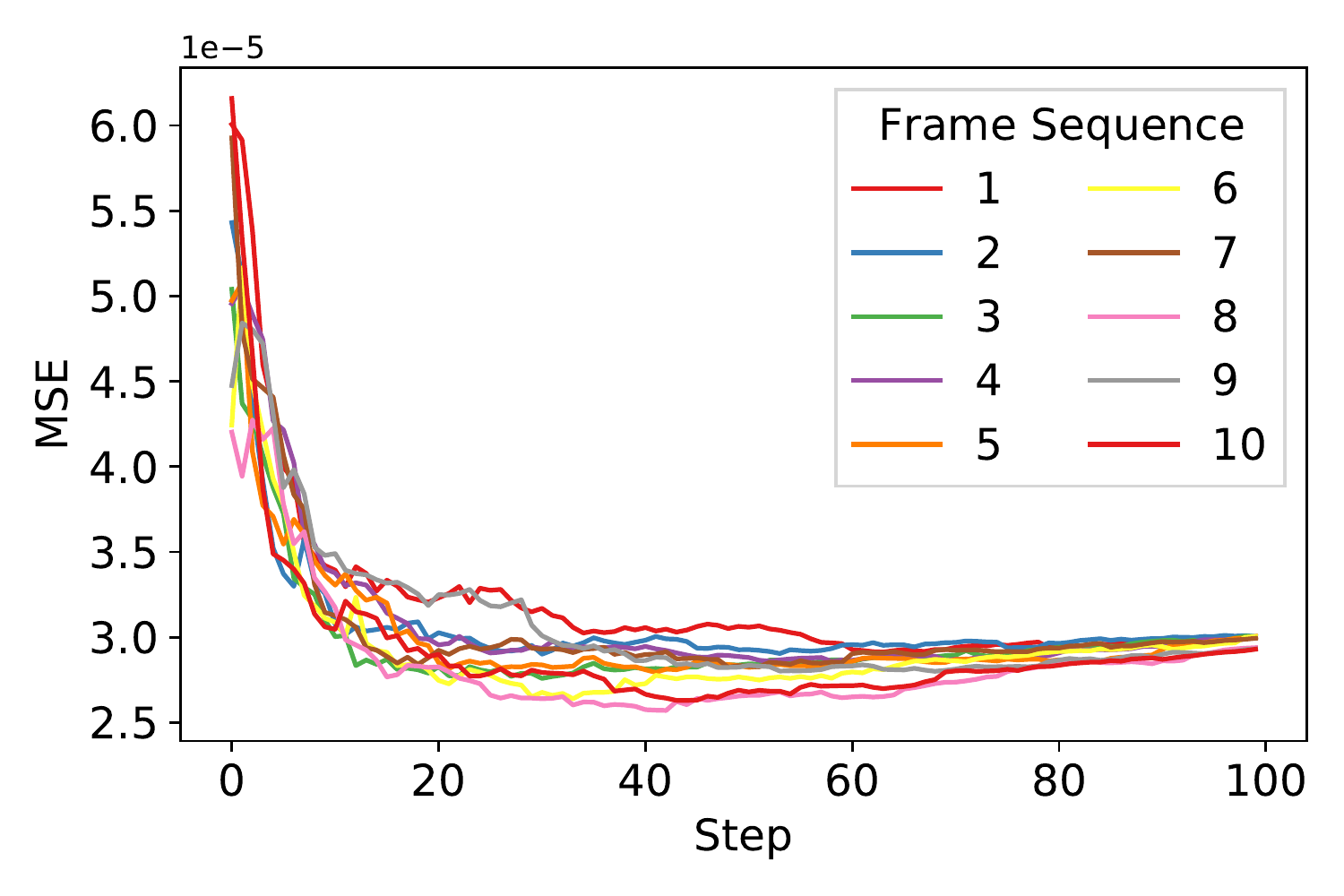} &
		\hspace{-9pt}\includegraphics[width=\sz\columnwidth]{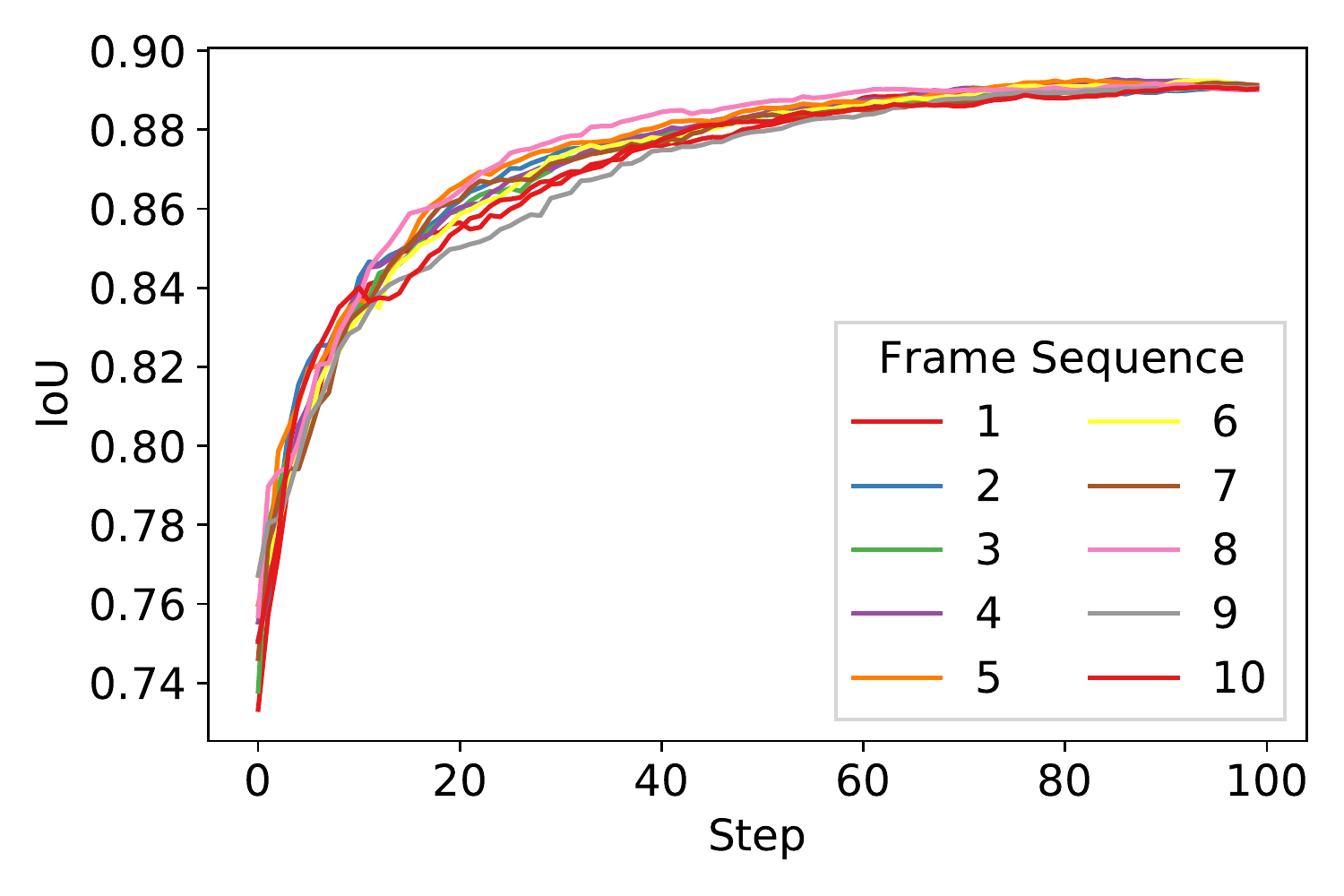} \\[-10pt]
	\end{tabular}
	\vspace{8pt}
	\caption{\textbf{Random frame order permutations.} The proposed method seems to be largely invariant to the frame integration order, since it always converges to the same result.}
	\label{fig:permutation-fusion}
	\vspace{-10pt}
\end{figure}

\boldparagraph{Frame Order Permutation.}
Our method does not leverage any temporal information from the camera trajectory apart from the previous fusion result.
This design choice allows to apply the method also to a broader class of scenarios (\eg Multi-View Stereo). 
Ideally, an online fusion method should be invariant to permutations of the fusion frame order. 
To verify this property, we evaluated the performance of our method in fusing the same set of frames in ten different random frame orders. 
Figure~\ref{fig:permutation-fusion} shows that our method converges to the same result for any frame order and thus seems to be invariant to frame order permutations.

\begin{table}[t]
	\centering
	\scriptsize
	\setlength{\tabcolsep}{3.5pt}
	\renewcommand{\arraystretch}{1.0}	
	\floatbox[\capbeside]{table}%
	{\caption{\textbf{Ablation Study.} We assess our method for different numbers of feature dimensions $\nFeatDim$. The performance saturates around $\nFeatDim = 8$. Note that $\nFeatDim = 1$ did not converge.}
	\label{tab:ablation_study}}%
	{\begin{tabular}{rrrrr}
		\toprule
		\textbf{N}  & \textbf{MSE}$\downarrow$  & \textbf{MAD}$\downarrow$   & \textbf{Acc.}$\uparrow$ & \textbf{IoU}$\uparrow$ \\
		& [e-05] & [e-02] & [\%] & [0,1] \\
		\midrule
		1  &  -   &   -  &  -    &  - \\
		2  & 9.45 & 0.64 & 94.67 & 0.717 \\
		4  & 4.03 & 0.30 & 97.51 & 0.863 \\
		8  & 3.99 & 0.29 & 97.46 & 0.862 \\
		16 & 3.91 & 0.29 & 97.50 & 0.863 \\
		\bottomrule
	\end{tabular}}
    \vspace{-12pt}  %
\end{table}
\boldparagraph{Feature Dimension.}
An important hyperparameter of our method is the feature dimension $\nFeatDim$.
Therefore, we show quantitative results for the reconstruction from noisy and outlier contaminated depth maps using varying $\nFeatDim$ in \tablename{}~\ref{tab:ablation_study}. 
We observe that a larger $\nFeatDim$ clearly improves the results, but the performance eventually saturates, which justifies our choice of $\nFeatDim = 8$ features throughout the paper.

\begin{figure}[h!]
	\centering
	\scriptsize
	\setlength{\tabcolsep}{0pt}
	\newcommand{\sz}{0.245}
	\begin{tabular}{cccc}
		\includegraphics[width=\sz\columnwidth]{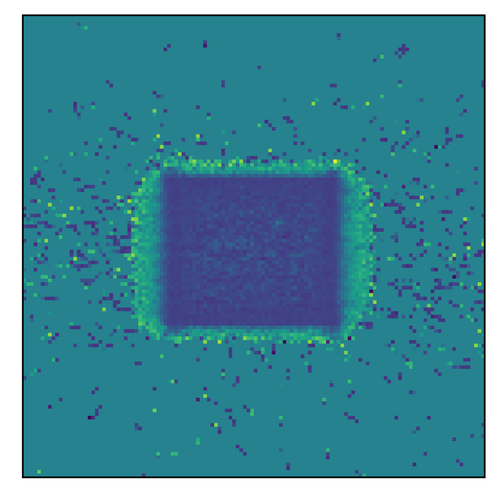} &
		\includegraphics[width=\sz\columnwidth]{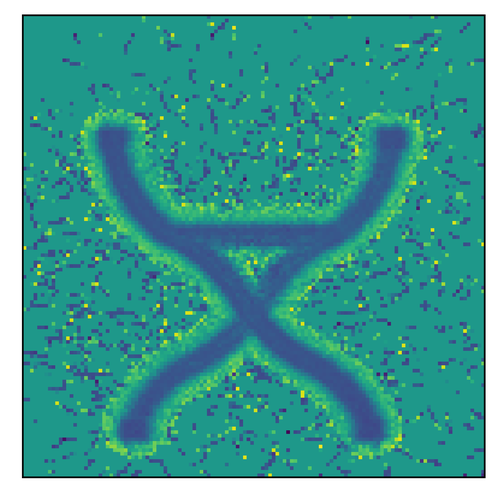} &
		\includegraphics[width=\sz\columnwidth]{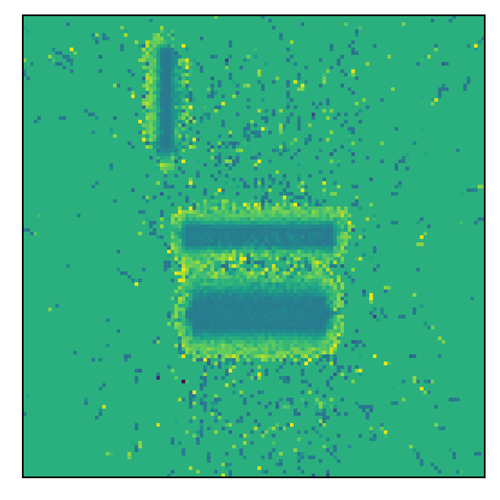} &
		\includegraphics[width=\sz\columnwidth]{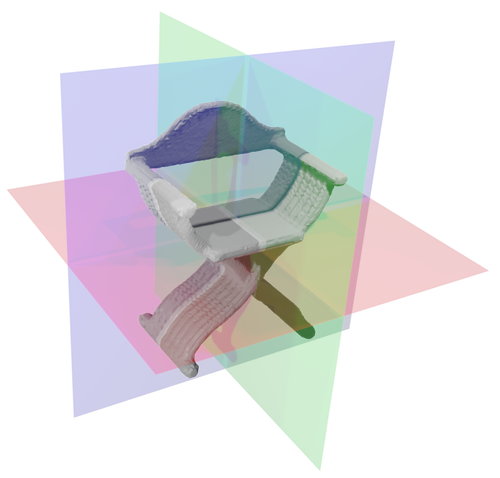} \\
		XY-Plane & XZ-Plane & YZ-Plane & Output Mesh\\[-5pt]
	\end{tabular}
	\vspace{8pt}
	\caption{\textbf{Visualization of our learned latent space encoding.} Our asynchronous fusion network integrates all measurements including outliers, but the translator effectively filters the outliers to generate a clean output mesh.}
	\label{fig:latent-space}
	\vspace{-10pt}
\end{figure}

\boldparagraph{Latent Space Visualization.}
In order to verify our hypothesis that the translator network mostly filters outliers, since the fusion network can hardly distinguish between first entries and outliers, we visualize the fused latent space in \figurename{}~\ref{fig:latent-space}.
While the translated end result is outlier free, the latent space clearly shows that the fusion network keeps track of most measurements.

\begin{figure*}[tb]
	\centering
	\scriptsize
	\setlength{\tabcolsep}{2pt}
	\newcommand{\sz}{0.252}
	\begin{tabular}{ccccc}
		\rotatebox{90}{\hspace{23pt}Lounge} &
		\includegraphics[height=2.5cm]{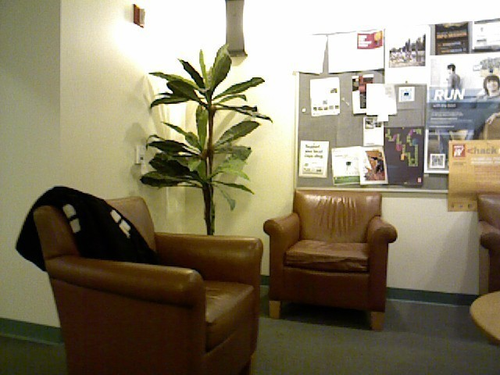} &
		\includegraphics[width=\sz\columnwidth]{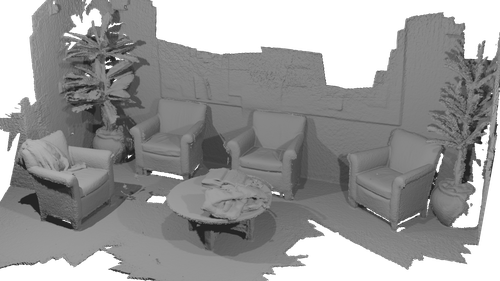} &
		\includegraphics[width=\sz\columnwidth]{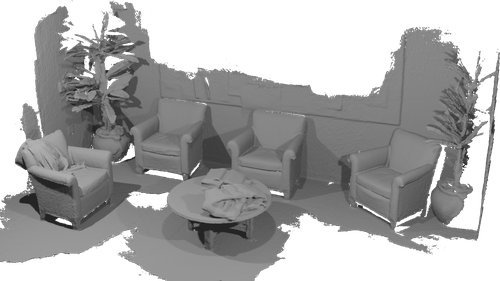} &
		\includegraphics[width=\sz\columnwidth]{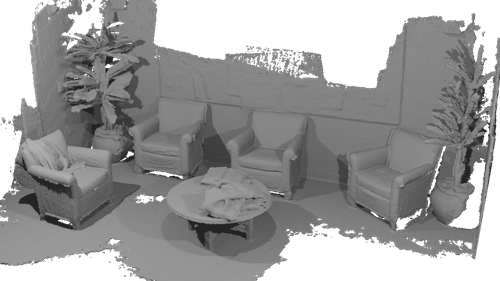} \\
		\rotatebox{90}{\hspace{18pt}Stone wall} &
		\includegraphics[height=2.5cm]{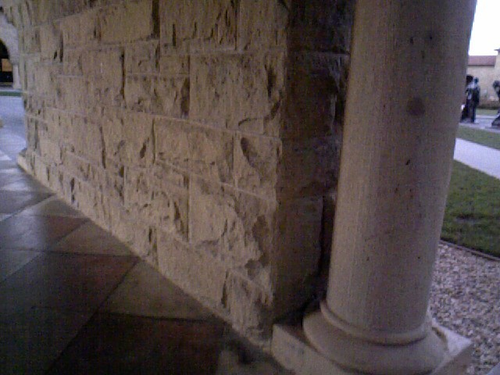} &
		\includegraphics[width=\sz\columnwidth]{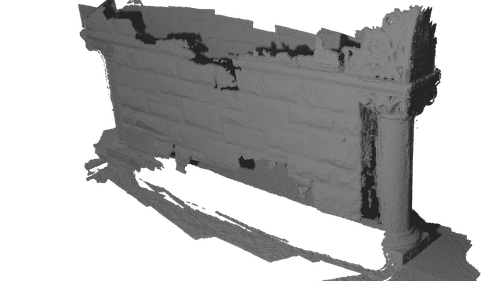} &
		\includegraphics[width=\sz\columnwidth]{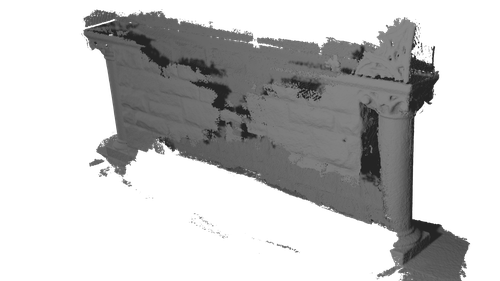} &
		\includegraphics[width=\sz\columnwidth]{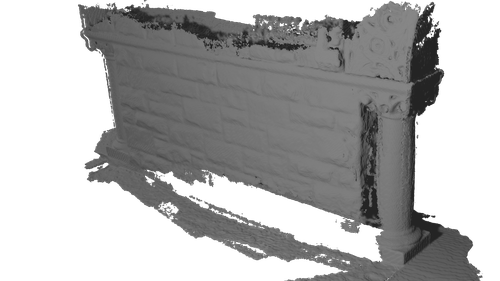} \\
		& Input Frame & TSDF Fusion~\cite{Curless-et-al-SIGGRAPH-1996} & RoutedFusion~\cite{Weder-et-al-CVPR-2020} & Ours\\[-5pt]
	\end{tabular}
	\caption{\textbf{Depth map fusion results on Scene3D~\cite{Zhou-et-al-SIGGRAPH-2013}}. Our method yields significantly better completeness than RoutedFusion~\cite{Weder-et-al-CVPR-2020} (see stonewall) and is on par with TSDF Fusion~\cite{Curless-et-al-SIGGRAPH-1996}. However, our method better removes noise and outlier artifacts (see lounge reconstruction).}
	\label{fig:scene3d-results}
\end{figure*}
\subsection{Real-World Data}
We also evaluate on real-world data and large-scale scenes to demonstrate scalability and generalization.

\boldparagraph{Scene3D Dataset.}
For real-world data evaluation, we use the lounge and stonewall scenes from the Scene3D dataset~\cite{Zhou-et-al-SIGGRAPH-2013}. 
For comparability to~\cite{Weder-et-al-CVPR-2020}, we only fuse every 10th frame from the trajectory using a model solely trained on synthetic ModelNet~\cite{Wu-et-al-CVPR-2015} data augmented with artificial noise and outliers.

\begin{table}[t]
	\centering
	\scriptsize
	\setlength{\tabcolsep}{1pt}
	\renewcommand{\arraystretch}{1}
	\newcommand{\pspace}{\hspace{5pt}}
	\scalebox{0.9}{
		\begin{tabular}{lrrc@{\pspace}rrc@{\pspace}rrc@{\pspace}rrc@{\pspace}rr}
			\toprule
			& \multicolumn{2}{c}{\textbf{Burghers}} && \multicolumn{2}{c}{\textbf{Stonewall}} && \multicolumn{2}{c}{\textbf{Lounge}} && \multicolumn{2}{c}{\textbf{Copyroom}} && \multicolumn{2}{c}{\textbf{Cactusgarden}} \\
			\cmidrule(lr){2-3} \cmidrule(lr){4-6} \cmidrule(lr){7-9} 
			\cmidrule(lr){10-12} \cmidrule(lr){13-15}
			\textbf{Method} & M.A. & M.C. && M.A. &  M.C. && M.A. & M.C. && M.A. &  M.C. && M.A. &  M.C. \\
			\hline \noalign{\vskip 1pt}
			TSDF Fusion~\cite{Curless-et-al-SIGGRAPH-1996}		& 21.01 & 22.58 && 17.67 & 21.16 && 21.88 & 26.31 && 39.56 & 42.57 && 18.91 & 18.63 \\
			RoutedFusion~\cite{Weder-et-al-CVPR-2020}	& 20.50 & 41.32 && 19.44 & 80.54 && 22.63 & 53.45 && 38.07 & 57.35 && 19.20 & 41.41 \\
			Ours 						& \bf 18.19 & \bf 18.88 && \bf 17.01 & \bf 20.27  && \bf 16.45 & \bf 17.96 && \bf 19.06 & \bf 20.25 && \bf 15.87 & \bf 16.96 \\[-2pt]
			\bottomrule
	\end{tabular}}
	\caption{\textbf{Quantitative evaluation on Scene3D~\cite{Zhou-et-al-SIGGRAPH-2013}.} Evaluated are mesh accuracy (M.A.) [mm] \& mesh completeness (M.C.) [mm].}
	\label{tab:quant-scene3d}
\end{table}
\begin{figure*}[htb!]
	\centering
	\scriptsize
	\setlength{\tabcolsep}{2pt}
	\newcommand{\sz}{0.188}
	\begin{tabular}{cccccc}
		\rotatebox{90}{\hspace{17pt}Caterpillar} & 
		\includegraphics[width=\sz\columnwidth]{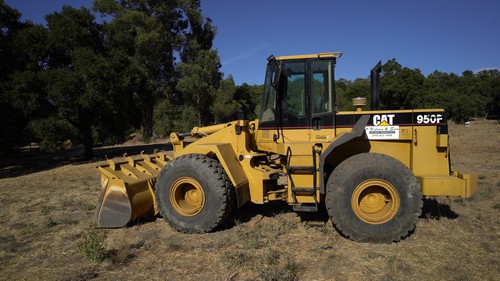} &
		\includegraphics[width=\sz\columnwidth]{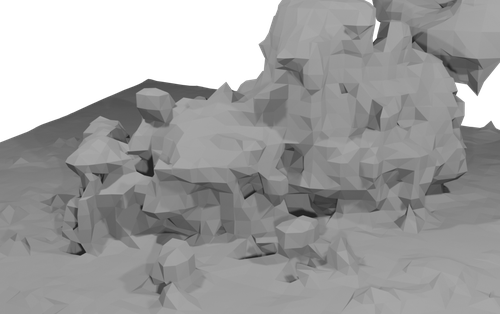} &
		\includegraphics[width=\sz\columnwidth]{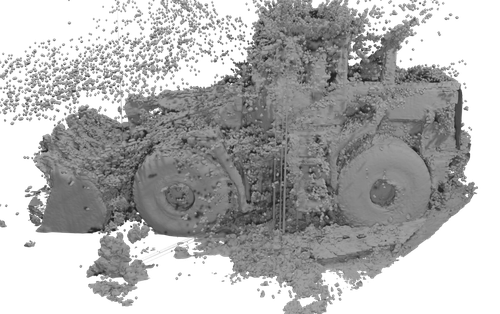} &
		\includegraphics[width=\sz\columnwidth]{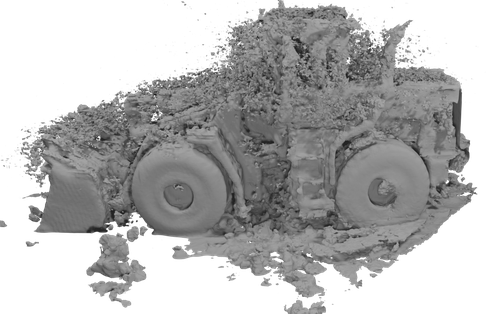} &
		\includegraphics[width=\sz\columnwidth]{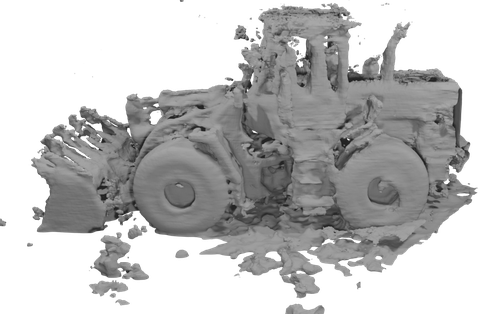} \\
		\rotatebox{90}{\hspace{20pt}Truck} & 
		\includegraphics[width=\sz\columnwidth]{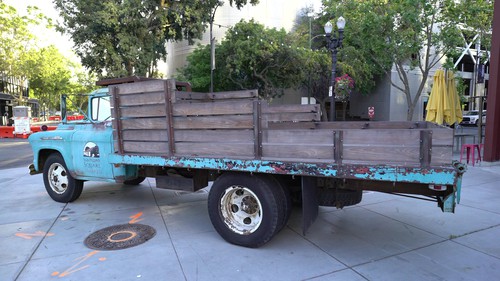} &
		\includegraphics[width=\sz\columnwidth]{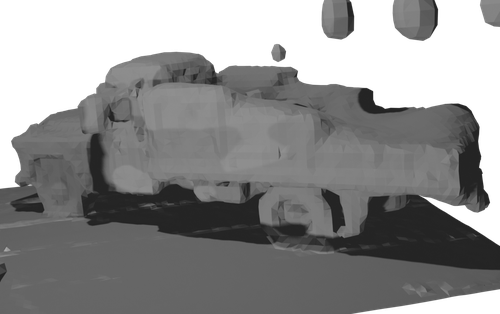} &
		\includegraphics[width=\sz\columnwidth]{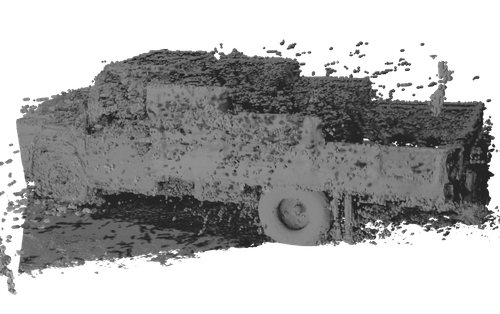} &
		\includegraphics[width=\sz\columnwidth]{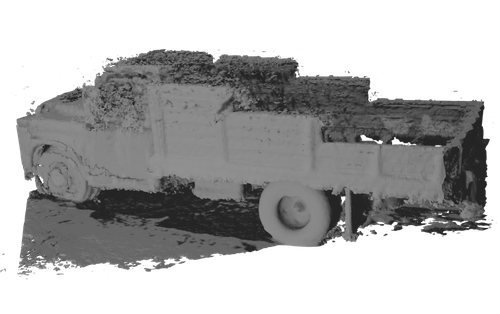} &
		\includegraphics[width=\sz\columnwidth]{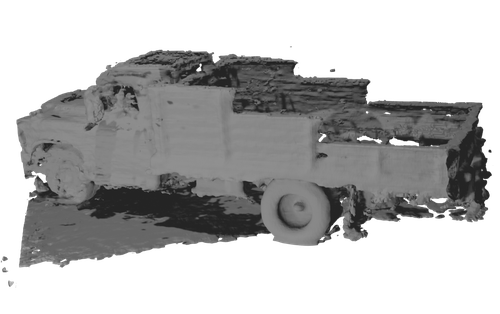} \\
		\rotatebox{90}{\hspace{23pt}M60} & 
		\includegraphics[width=\sz\columnwidth]{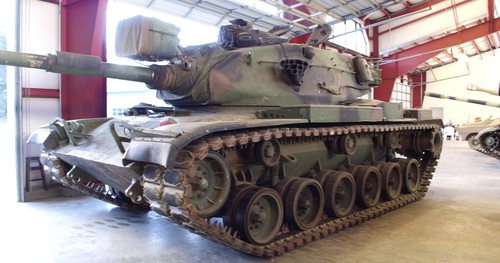} &
		\includegraphics[width=\sz\columnwidth]{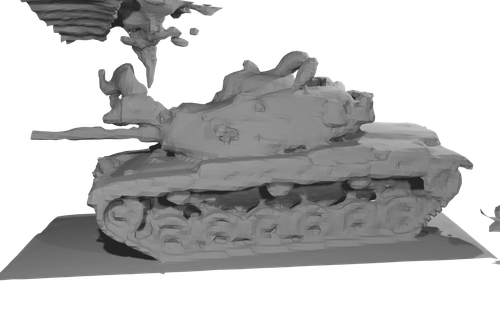} &
		\includegraphics[width=\sz\columnwidth]{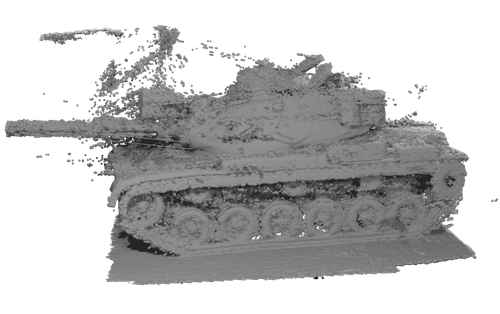} &
		\includegraphics[width=\sz\columnwidth]{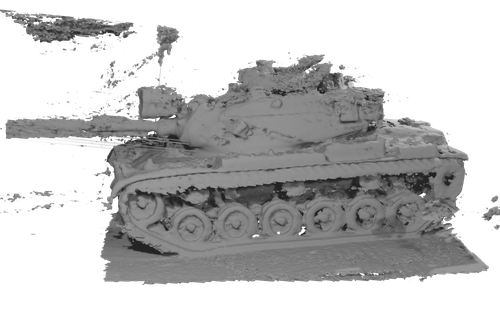} &
		\includegraphics[width=\sz\columnwidth]{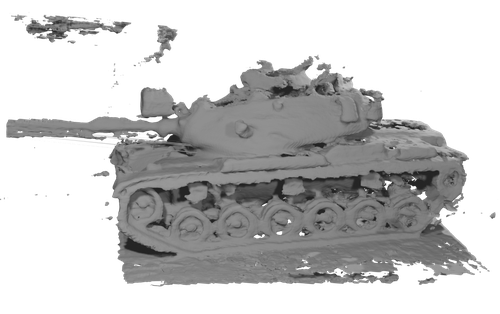} \\
		& 
		Input Frame & PSR~\cite{Kazhdan-Hoppe-TOG-2013} & TSDF Fusion~\cite{Curless-et-al-SIGGRAPH-1996} & RoutedFusion~\cite{Weder-et-al-CVPR-2020} & Ours\\[-5pt]
	\end{tabular}
	\caption{\textbf{Results on Tanks and Temples~\cite{Knapitsch-et-al-TOG-2017}}. Our method significantly reduces the number of outliers compared to the other methods without using any outlier filtering heuristic and solely learning it from data. We especially highlight the results on the caterpillar scene, where our proposed method filters most outliers while the reconstructions of competing methods are heavily cluttered with outliers.}
	\label{fig:tankstemples-results}
\end{figure*}
In \figurename{}~\ref{fig:scene3d-results}, we present qualitative results of reconstructions from real-world depth maps compared to RoutedFusion~\cite{Weder-et-al-CVPR-2020} and TSDF Fusion~\cite{Curless-et-al-SIGGRAPH-1996}.
We note that our method reconstructs the scene with significantly higher completeness than~\cite{Weder-et-al-CVPR-2020}.
This is due to our learned translation from feature to TSDF space, which allows to better handle noise artifacts and outliers without the need for hand-tuned, heuristic post-filtering.
Further, we show improved outlier and noise artefact removal compared to TSDF Fusion~\cite{Curless-et-al-SIGGRAPH-1996} while being on par with respect to completeness.
These results are also quantitatively shown in~\tablename{}~\ref{tab:quant-scene3d}.

\boldparagraph{Tanks and Temples Dataset.}
In order to demonstrate our methods outlier handling capability, we also run experiments on the Tanks and Temples dataset~\cite{Knapitsch-et-al-TOG-2017}. 
We computed stereo depth maps using COLMAP~\cite{Schoenberger-et-al-CVPR-2016,Schoenberger-et-al-ECCV-2016} and fused this data using our method, PSR~\cite{Kazhdan-Hoppe-TOG-2013}, TSDF Fusion~\cite{Curless-et-al-SIGGRAPH-1996}, and RoutedFusion~\cite{Weder-et-al-CVPR-2020}. 
To demonstrate the easy applicability to new datasets in scenarios with limited ground-truth, we train our method on one single scene (Ignatius) from the Tanks and Temples training set. 
We reconstructed a dense mesh using Poisson Surface Reconstruction (PSR)~\cite{Kazhdan-Hoppe-TOG-2013}, rendered artificial depth maps, and used TSDF Fusion to generate a ground-truth SDF.
Then, we used this ground-truth to train the fusion of stereo depth maps.
\figurename{}~\ref{fig:tankstemples-results} shows the reconstructions of the unseen Caterpillar, Truck, and M60 scene from~\cite{Knapitsch-et-al-TOG-2017}. 
Our proposed method significantly reduces the amount of outliers in the scene across all models. 
While~\cite{Weder-et-al-CVPR-2020} shows comparable results on some scenes, it is heavily dependent on its outlier post-filter, which fails as soon as there are too many outliers in the scene (see also \figurename{}~\ref{fig:teaser}). 
Further, they also pre-process the depth maps using a 2D denoising network while our network uses the raw depth maps.

\boldparagraph{Limitations.} While our pipeline shows excellent generalization capabilities (\eg generalizing from a single MVS training scene), it is biased to the number of observations integrated during training leading to less complete results on some test scene parts with very few observations. 
However, this issue can be overcome by a more diverse set of training sequences with different number of observations.
\section{Conclusion}

We presented a novel approach to online depth map fusion with real-time capability.
The key idea is to perform the fusion operation in a learned latent space that allows to encode additional information about undesired outliers and super-resolution complex shape information.
The separation of scene representations for fusion and final output allows for an end-to-end trainable post-filtering as a translator network, which takes the latent scene encoding and decodes it into a standard TSDF representation.
Our experiments on various synthetic and real-world datasets demonstrate superior reconstruction results, especially in the presence of large amounts of noise and outliers.

\noindent
\begin{minipage}{\columnwidth}
	\vspace{0mm}
	\footnotesize
	\noindent
	\textbf{Acknowledgments.}~
	We thank Akihito Seki from Toshiba Japan for insightful discussions and valuable comments.
This research was supported by Toshiba and Innosuisse funding (Grant No. 34475.1 IP-ICT).
\end{minipage}

\appendix

\section{Evaluation Metrics}

In order to compare our method to state-of-the-art learning-based methods and standard TSDF fusion, we compute the following six metrics on reconstructions:

\boldparagraph{Mean Squared Error (MSE) and Mean Absolute Distance (MAD).}
The mean squared error measures the reconstruction error on the TSDF field by penalizing large surface deviations and outliers.
The mean absolute distance is also computed on the TSDF grid.
However, it mainly quantifies the performance on reconstructing fine geometric details.

\boldparagraph{Accuracy (Acc.), F1 Score, Intersection-over-Union (IoU):}
The accuracy is computed over the occupancy obtained from the sign of the TSDF grid.
We also report the F1 score, which is the harmonic mean of precision and recall. 
By measuring both, completeness and accuracy, it is a more holistic metric for quantifying the performance of a reconstruction method.
Moreover, we measure the IoU on the occupancy grid.
The IoU especially quantifies artifacts typically encountered in reconstructions from noisy depth maps, such as surface and corner thickening and the vanishing of fine geometric details.

\boldparagraph{Mesh Completeness (M.C.) and Accuracy (M.A.)}
We compute the completeness using the evaluation pipeline from~\cite{Stutz-et-al-CVPR-2018}.
The completeness describes the distance from points sampled on the ground-truth mesh to the closest point on the reconstructed mesh.
Vice-versa, the accuracy computes the distance from points sampled on the reconstructed mesh to the closest point on the ground-truth mesh.

\section{Reproducibility}

For reproducibility, our source code will be made publicly available upon publication.
We further present more details of our fusion pipeline in the following.

\subsection{Details on Pipeline Architecture}
Our method consists of four neural network components that are used for \textbf{(i)} sub-volume \textbf{extraction} of the global canonical feature volume, \textbf{(ii) fusion} of previously fused feature with a new depth map , \textbf{(iii) integration} of the fused updates back into the global feature volume, and \textbf{(iv) translation} from the latent feature space to TSDF and occupancy. %

\boldparagraph{(i) Extraction Layer.}
In the extraction layer, we extract the current state of the global feature volume into a view-dependent canonical feature volume defined by the camera parameters of the current measurement. 
In a first step, we un-project all depth pixels into the global feature grid using the camera parameters:
\begin{equation}
p_{XYZ} = \begin{bmatrix} R & t\end{bmatrix}^{-1}\begin{bmatrix}K^{-1}p & 1\end{bmatrix}
\end{equation}

where $p_{XYZ}$ are the coordinates of the un-projected point in world coordinates and $p = (p_x, p_y, d)^T$ are the pixel coordinates with its corresponding depth measurements. 
In a second step, we sample points around $p_{XYZ}$ in a window centered at $p_{XYZ}$ and aligned with the direction of the viewing ray. 
This procedure is inspired by the extraction step used in~\cite{Weder-et-al-CVPR-2020}.
Finally, we convert the coordinates of each sampled point to grid coordinates and extract the current feature state using nearest-neighbor interpolation.

\boldparagraph{(ii) Feature Fusion Network.}
The feature fusion network consists of three components: Feature Encoder, Feature Decoder, and Feature Normalization, which are detailed in the following.
\begin{enumerate}%
\item \textbf{Feature Encoder:} The feature encoder is built from four network blocks each consisting of the following modules: 1) a 2D convolution having kernel size of 3 and zero padding reducing the number of input channels, 2) a layer normalization, 3) tanh activation, 4) again a 2D convolution having kernel size of 3 and zero padding but without reducing the channels, 5) layer normalization, and 6) tanh activation.  The ouput of each block is concatenated with its input and passed to the next block.
\item \textbf{Feature Decoder:}  The feature decoder also consists of four neural blocks, of which each has the same design as the neural blocks in the feature encoder.  However, instead of having a kernel-size of three, the convolutional layers in the feature decoder have a kernel-size of one.  The motivation behind this choice is that the encoder has already encoded enough neighboring information and, therefore, the decoder predicts the feature updates based on the encoded information for each ray separately.
\item \textbf{Feature Normalization:}  After predicting the feature updates for each position in the local, view-dependent feature volume, we normalize each feature vector.  This normalization prevents the feature values from becoming too large and, therefore, it improves the pipeline's capability to update the scene.
\end{enumerate}

\boldparagraph{(iii) Integration Layer.} 
In the integration layer, we integrate the predicted feature updates from the fusion network back into the global feature volume. 
Therefore, we aggregate all updates that are mapped to the same global feature grid location using the correspondence given by the camera parameters. 
Then, we use an average pooling to combine multiple correspondences to the same grid location. 
Finally, we update the feature volume by using a running average update similar to~\cite{Curless-et-al-SIGGRAPH-1996}.

\boldparagraph{(iv) Feature Translation Network.}
The feature translation network renders the output modalities (TSDF and occupancy) from the latent feature representation for a specific query point $\nQueryPoint_i$.
It consists of three components: a neighborhood interpolator, a translation MLP, and two network heads predicting the output modalities.
\begin{enumerate}%
\item \textbf{Neighborhood Interpolator:} The neighborhood interpolator encodes information from the neighboring feature vector into a single feature vector. 
Therefore, all neighboring feature vectors are concatenated and passed through a single linear layer followed by tanh activation. 
The output has the same dimension as one single feature vector.
\item \textbf{Translation MLP:} The output of the neighborhood interpolator is concatenated with the query point feature vector $\nFeatGrid_t(\nQueryPoint_i)$ and passed through the translation MLP. 
The translation MLP is built from four linear layers interleaved with tanh activations. 
During training, the output of each layer is further passed through a channel-wise dropout layer with dropout probability $p=0.2$. 
The first layer has 32 output channels, the second layer has 16 output channels, and the third and fourth layer have each 8 output channels. 
The output of each layer is concatenated with the query point feature vector. 
The output of the final layer is then passed to the two network output heads.
\item \textbf{Network Output Heads:} The translation network has two network output heads. 
Each head takes the output of the MLP as an input and predicts a translation modality. 
The TSDF head predicts the TSDF using a single linear layer outputting one channel that is followed by a tanh activation. 
The output of the activation is further scaled by the truncation band of the ground-truth TSDF ($0.04$) to map it into the correct value range. 
The occupancy head is also passed through a linear layer but activated using a sigmoid activation to map into a unit interval.
\end{enumerate}

\subsection{Details on Hyperparameters}
\tablename~\ref{tab:hyperparameters} summarizes the choice of all hyperparameters that we used for all experiments in our work.
\begin{table}[h]
	\centering
	\caption{Network hyperparameters}
	\label{tab:hyperparameters}
	\begin{tabular}{lr}
		\toprule
		\multicolumn{2}{c}{\textbf{Optimizer}} \\
		\midrule
		Name & ADAM \\
		$\beta_1$ & $0.9\phantom{99}$ \\
		$\beta_2$ & $0.999$ \\
		$\epsilon$ & $1.e-08$ \\
		\midrule
		\multicolumn{2}{c}{\textbf{Learning Rate Scheduling}} \\
		\midrule
		Initial Learning Rate & $0.01\phantom{8}$ \\
		Decay & $0.998$ \\
		\midrule
		\multicolumn{2}{c}{\textbf{Loss Weights}} \\
		\midrule
		$\lambda_1$ & $1.0\phantom{00}$\\
		$\lambda_2$ & $10.0\phantom{00}$\\
		$\lambda_{\nOcc}$ & $0.01\phantom{0}$ \\
		$\lambda_{\nFeatGrid}$ & $0.05\phantom{0}$ \\
		\bottomrule
	\end{tabular}
\end{table}

\section{Qualtitative Results}
In this section, we present more qualitative results to demonstrate the performance of our method.

\subsection{Synthetic Data}

\begin{figure}[th]
	\centering
	\includegraphics[width=\columnwidth]{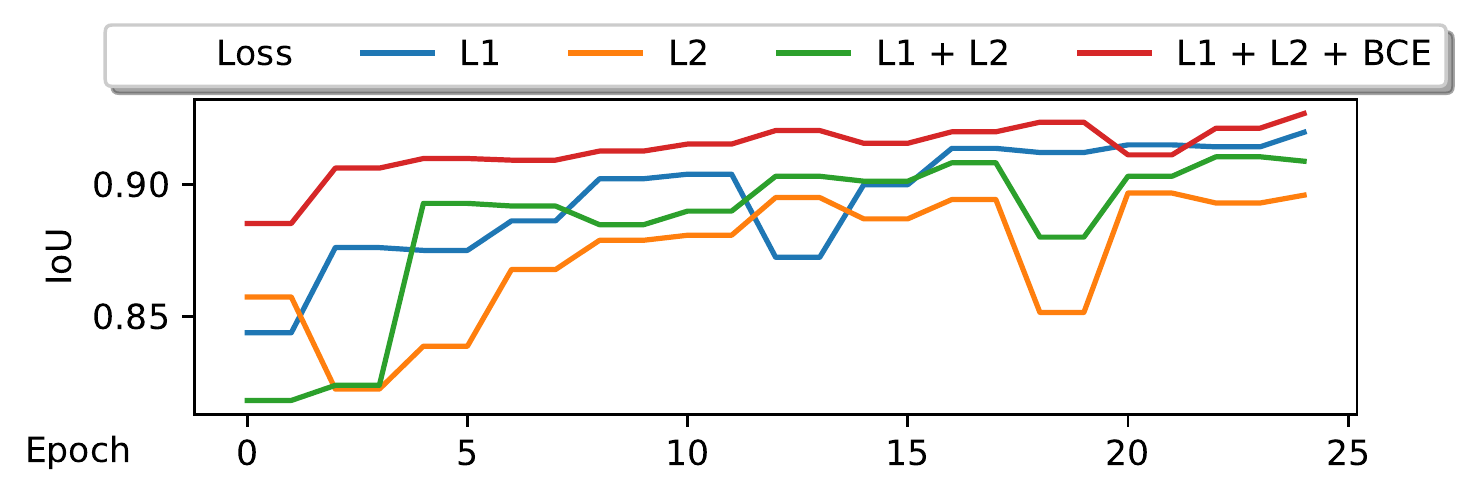}
	\vspace{-18pt}
	\caption{\textbf{Loss Ablation.} All losses combined yield best results.}
	\label{fig:loss-ablation}
	\vspace{-7pt}
\end{figure}

\begin{figure*}[!htb]
	\centering
	\scriptsize
	\setlength{\tabcolsep}{2pt}
	\newcommand{\sz}{0.132}
	\begin{tabular}{ccccccc}
		& \multicolumn{3}{c}{\textbf{Mesh Reconstruction}} & \multicolumn{3}{c}{\textbf{Outlier Projection}} \\
		\raisebox{0.1\height }{\rotatebox{90}{TSDF Fusion~\cite{Curless-et-al-SIGGRAPH-1996}}} &
		\includegraphics[width=\sz\textwidth]{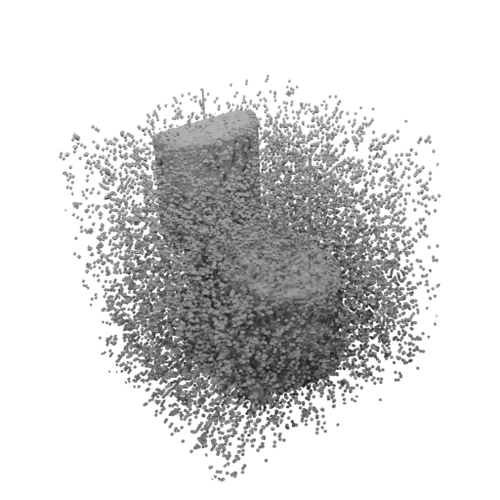} &
		\includegraphics[width=\sz\textwidth]{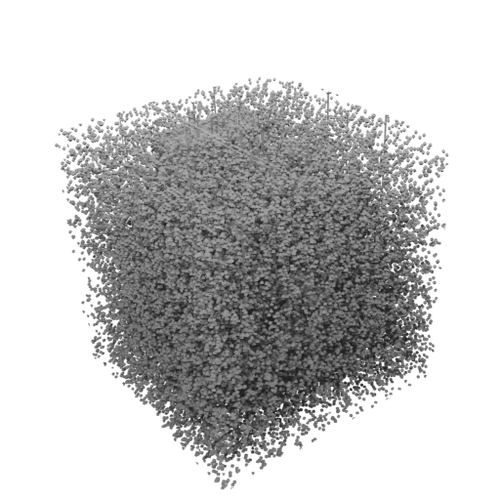} &
		\includegraphics[width=\sz\textwidth]{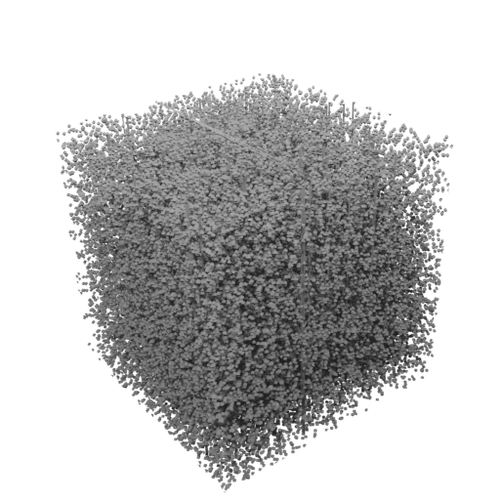} &
		\includegraphics[width=\sz\textwidth]{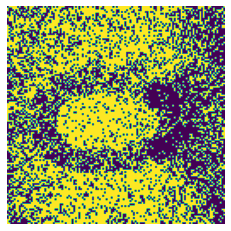} &
		\includegraphics[width=\sz\textwidth]{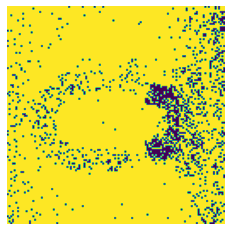} &
		\includegraphics[width=\sz\textwidth]{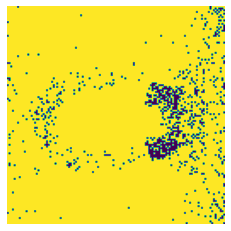} \\
		\raisebox{0.1\height }{\rotatebox{90}{RoutedFusion~\cite{Weder-et-al-CVPR-2020}}} &
		\includegraphics[width=\sz\textwidth]{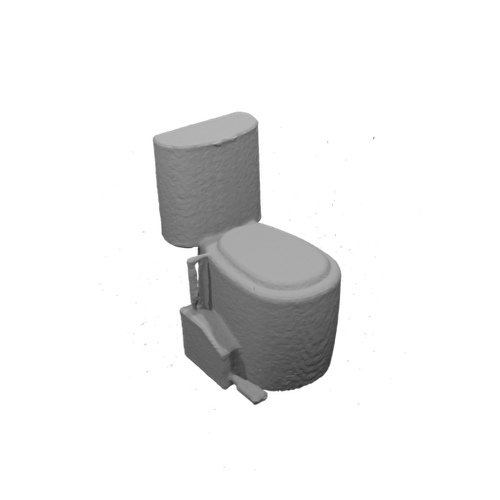} &
		\includegraphics[width=\sz\textwidth]{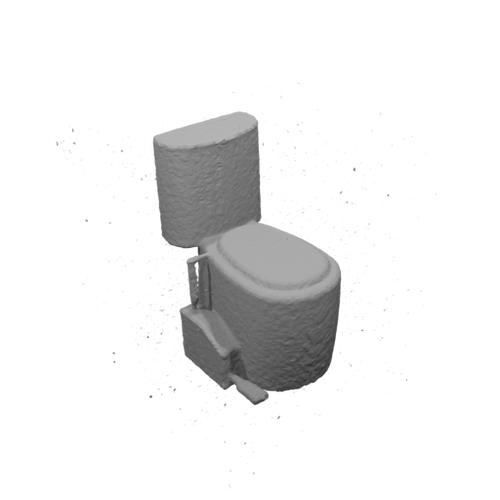} &
		\includegraphics[width=\sz\textwidth]{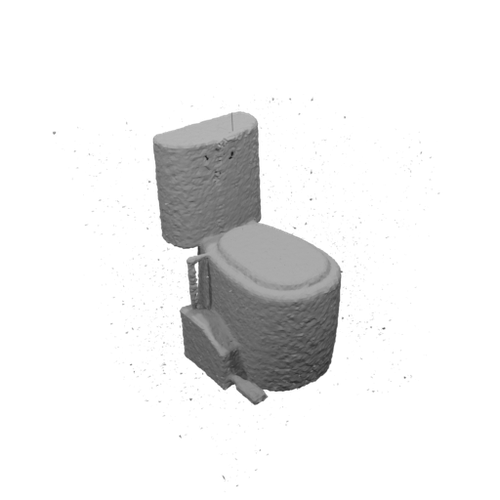} &
		\includegraphics[width=\sz\textwidth]{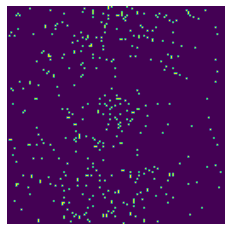} &
		\includegraphics[width=\sz\textwidth]{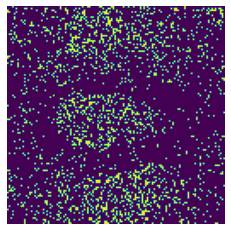} &
		\includegraphics[width=\sz\textwidth]{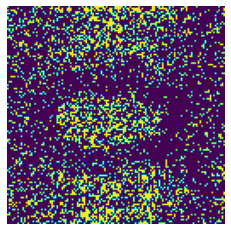} \\
		\raisebox{1.7\height }{\rotatebox{90}{Ours}} &
		\includegraphics[width=\sz\textwidth]{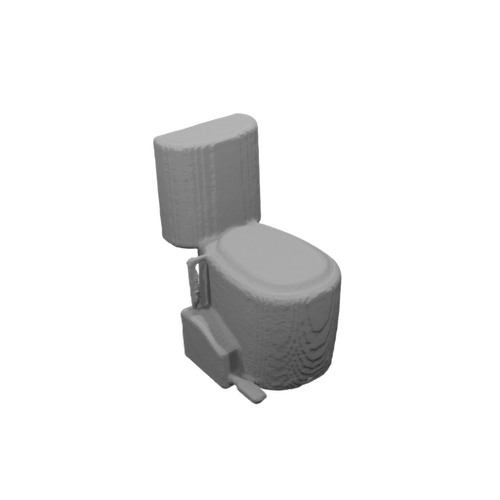} &
		\includegraphics[width=\sz\textwidth]{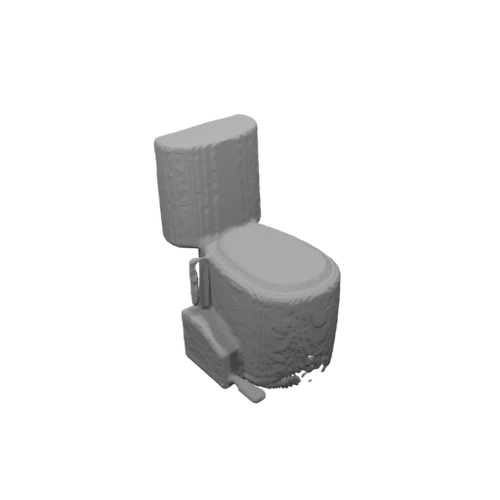} &
		\includegraphics[width=\sz\textwidth]{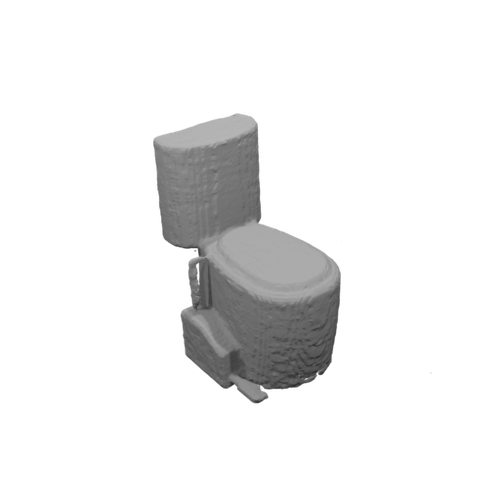} &
		\includegraphics[width=\sz\textwidth]{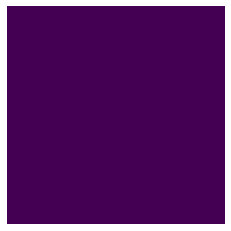} &
		\includegraphics[width=\sz\textwidth]{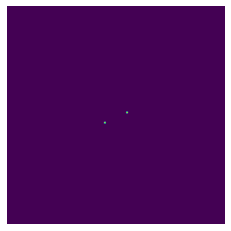} &
		\includegraphics[width=\sz\textwidth]{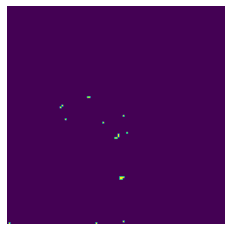} \\
		\raisebox{0.1\height }{\rotatebox{90}{TSDF Fusion~\cite{Curless-et-al-SIGGRAPH-1996}}} &
		\includegraphics[width=\sz\textwidth]{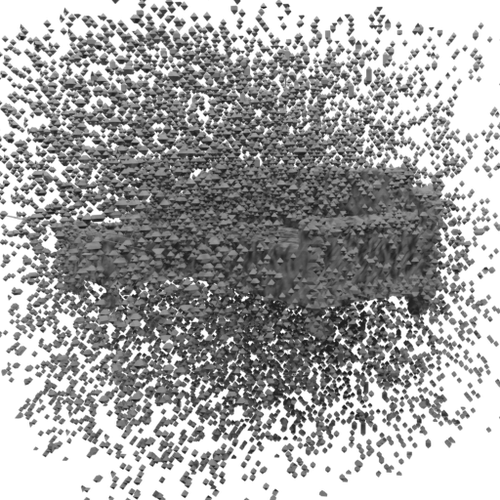} &
		\includegraphics[width=\sz\textwidth]{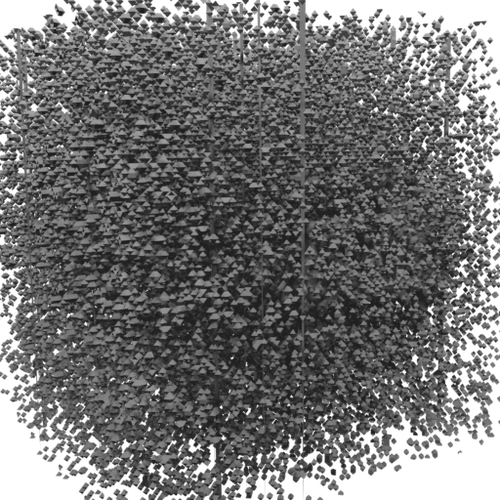} &
		\includegraphics[width=\sz\textwidth]{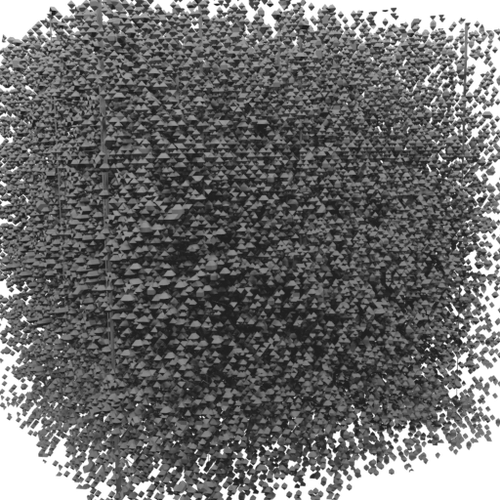} &
		\includegraphics[width=\sz\textwidth]{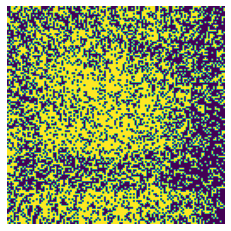} &
		\includegraphics[width=\sz\textwidth]{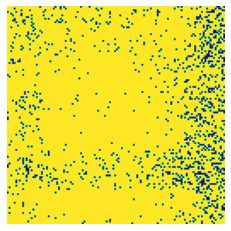} &
		\includegraphics[width=\sz\textwidth]{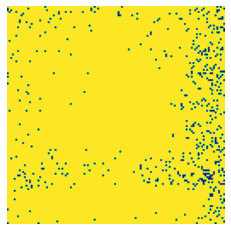} \\
		\raisebox{0.1\height }{\rotatebox{90}{RoutedFusion~\cite{Weder-et-al-CVPR-2020}}} &
		\includegraphics[width=\sz\textwidth]{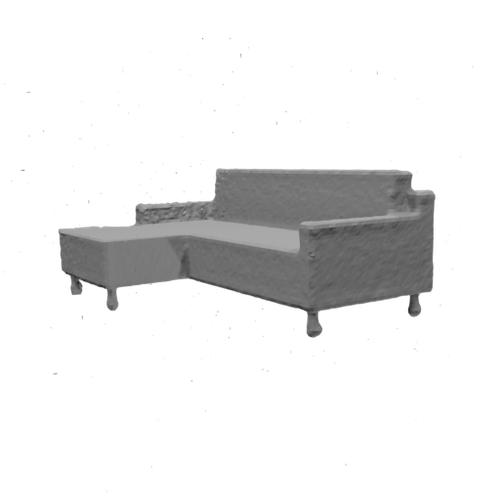} &
		\includegraphics[width=\sz\textwidth]{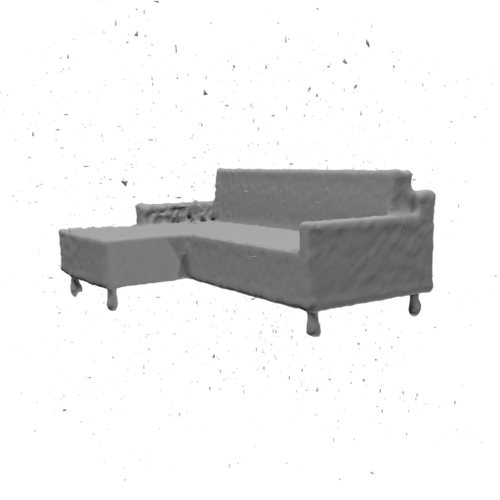} &
		\includegraphics[width=\sz\textwidth]{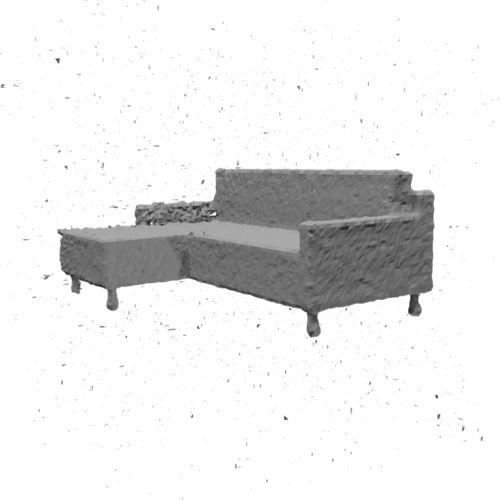} &
		\includegraphics[width=\sz\textwidth]{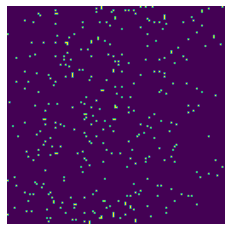} &
		\includegraphics[width=\sz\textwidth]{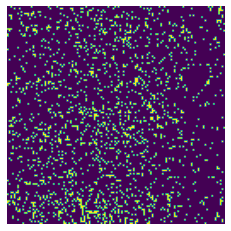} &
		\includegraphics[width=\sz\textwidth]{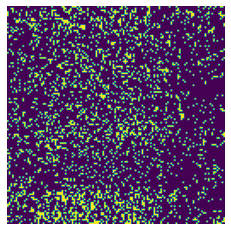} \\
		\raisebox{1.7\height }{\rotatebox{90}{Ours}} &
		\includegraphics[width=\sz\textwidth]{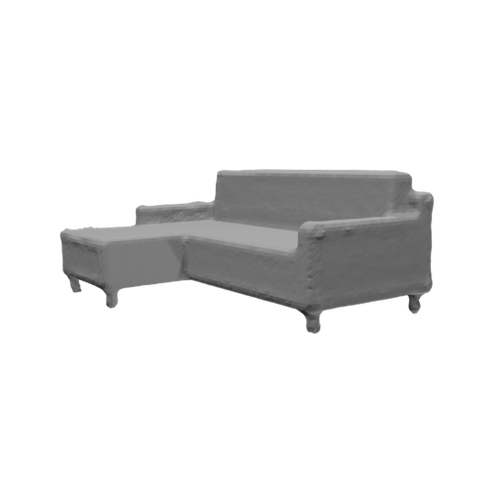} &
		\includegraphics[width=\sz\textwidth]{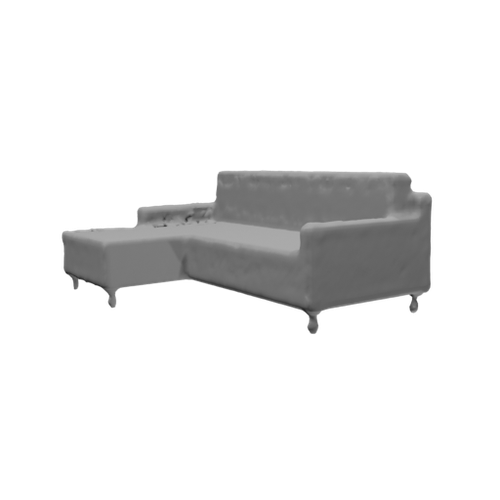} &
		\includegraphics[width=\sz\textwidth]{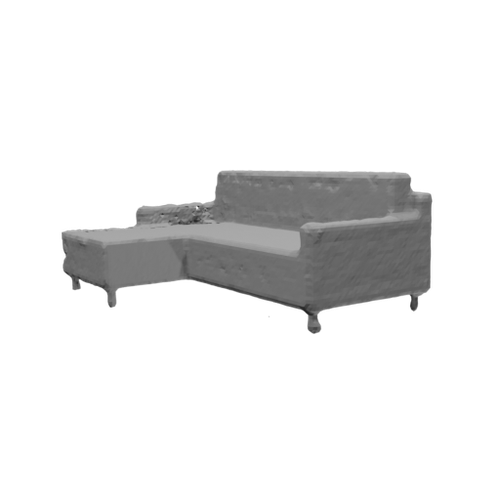} &
		\includegraphics[width=\sz\textwidth]{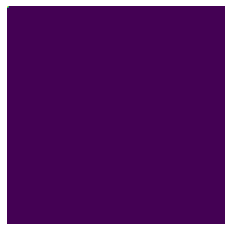} &
		\includegraphics[width=\sz\textwidth]{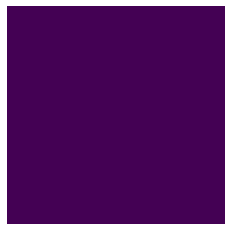} &
		\includegraphics[width=\sz\textwidth]{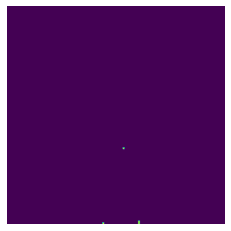} \\
		\raisebox{0.1\height }{\rotatebox{90}{TSDF Fusion~\cite{Curless-et-al-SIGGRAPH-1996}}} &
		\includegraphics[width=\sz\textwidth]{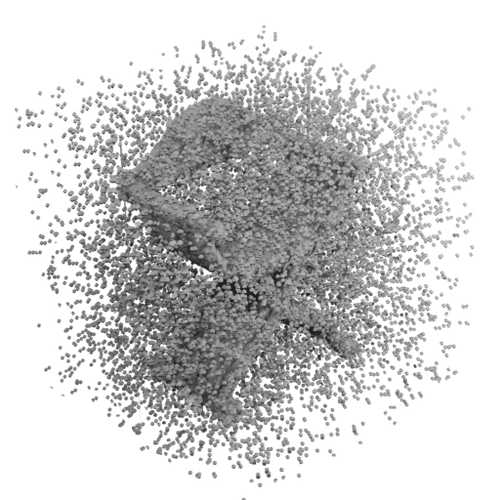} &
		\includegraphics[width=\sz\textwidth]{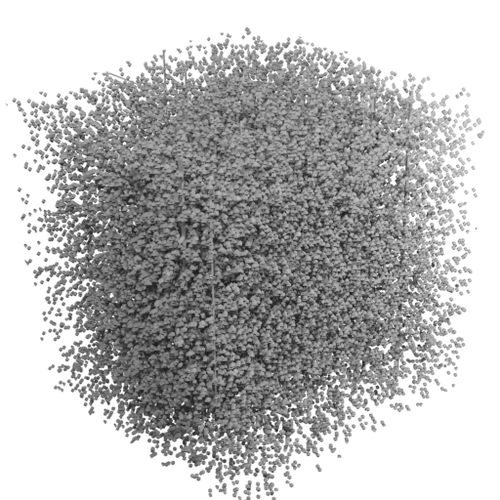} &
		\includegraphics[width=\sz\textwidth]{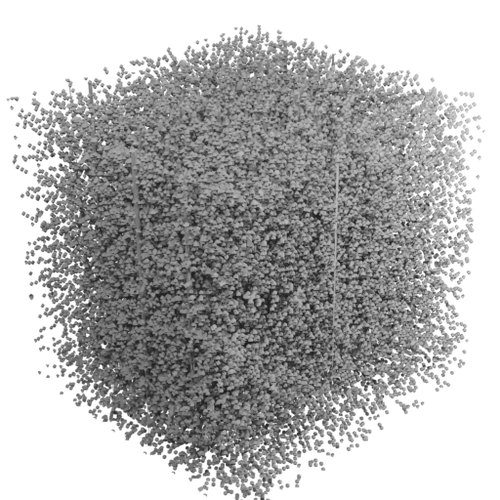} &
		\includegraphics[width=\sz\textwidth]{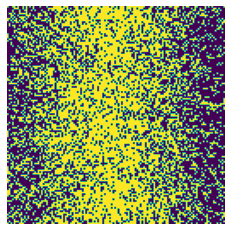} &
		\includegraphics[width=\sz\textwidth]{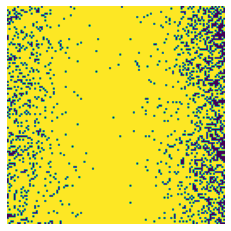} &
		\includegraphics[width=\sz\textwidth]{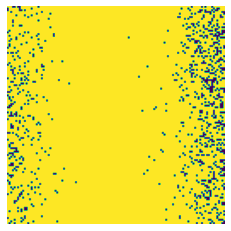} \\
		\raisebox{0.1\height }{\rotatebox{90}{RoutedFusion~\cite{Weder-et-al-CVPR-2020}}} &
		\includegraphics[width=\sz\textwidth]{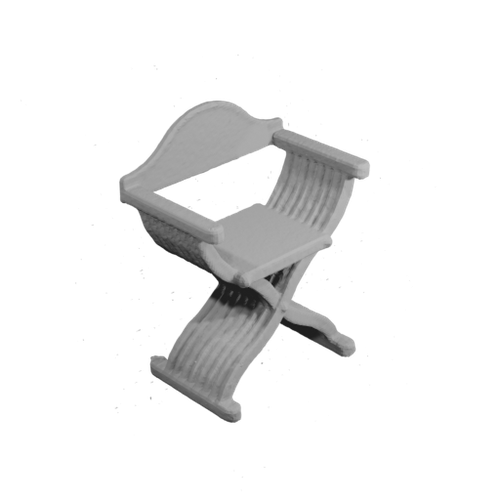} &
		\includegraphics[width=\sz\textwidth]{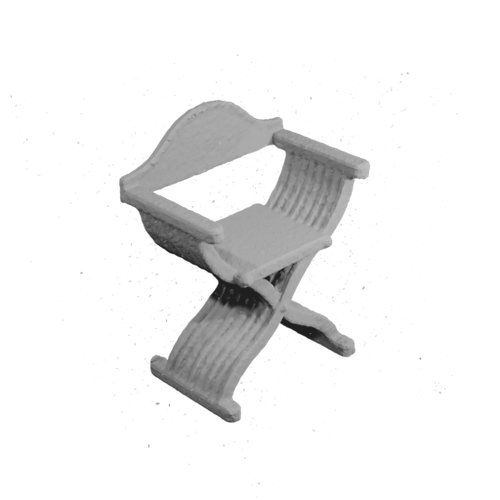} &
		\includegraphics[width=\sz\textwidth]{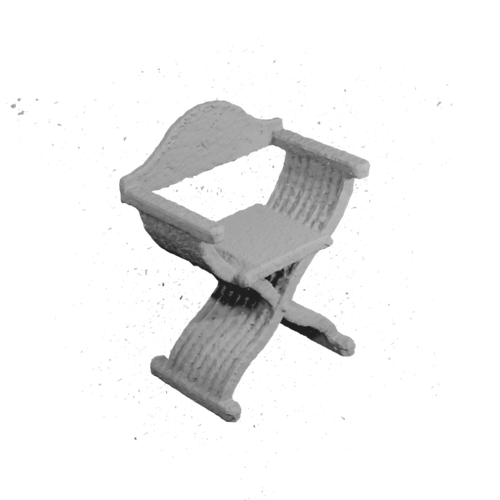} &
		\includegraphics[width=\sz\textwidth]{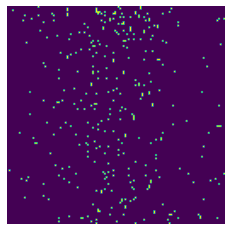} &
		\includegraphics[width=\sz\textwidth]{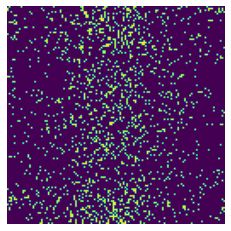} &
		\includegraphics[width=\sz\textwidth]{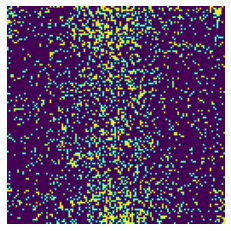} \\
		\raisebox{1.7\height }{\rotatebox{90}{Ours}} &
		\includegraphics[width=\sz\textwidth]{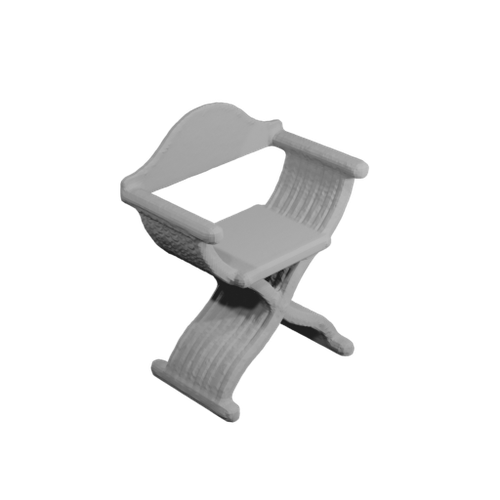} &
		\includegraphics[width=\sz\textwidth]{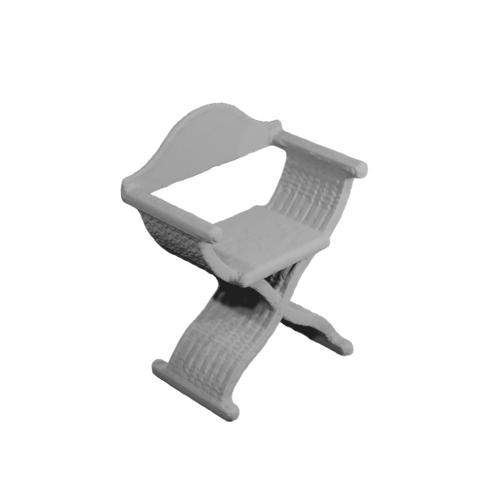} &
		\includegraphics[width=\sz\textwidth]{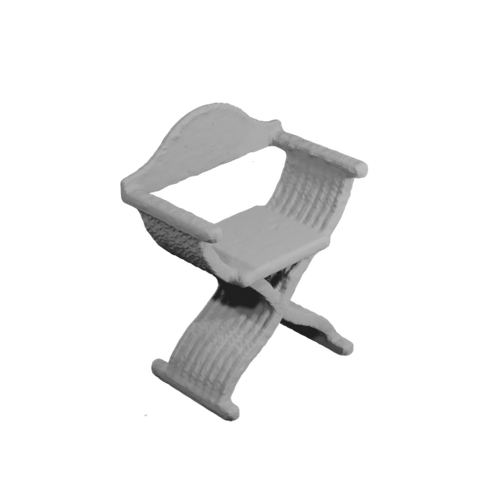} &
		\includegraphics[width=\sz\textwidth]{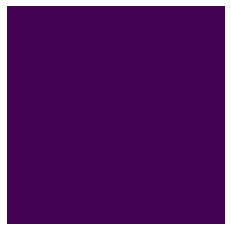} &
		\includegraphics[width=\sz\textwidth]{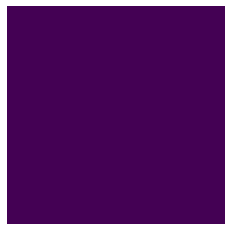} &
		\includegraphics[width=\sz\textwidth]{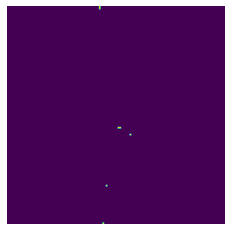} \\
		& 0.01 & 0.05 & 0.1 & 0.01 & 0.05 & 0.1 \\[-8pt]
	\end{tabular}
	\caption{\textbf{More qualitative results for different outlier fractions on ModelNet~\cite{Wu-et-al-CVPR-2015} examples.} Our method consistently removes more outliers than existing depth map fusion methods. Even for large outlier fractions, our method successfully filters almost all of them.}
	\label{fig:outlier-qualitative}
\end{figure*}

In \figurename~\ref{fig:outlier-qualitative}, we show additional qualitative results for the outlier robustness experiment that we presented in the main paper.
Our method is consistently better in filtering outliers than existing methods that have no outlier filtering or filter outliers heuristically.

\subsection{Real-World Data}
\begin{figure*}[!htb]
	\centering
	\scriptsize
	\newcommand{\sz}{0.42}
	\setlength{\tabcolsep}{3pt}
	\begin{tabular}{c@{\hspace{2pt}}ccc}
	\multirow{1}{*}[60pt]{\rotatebox{90}{Cactus garden}} & &
	\begin{tikzpicture}
	\node[anchor=south west,inner sep=0] at (0,0) {\includegraphics[width=\sz\textwidth]{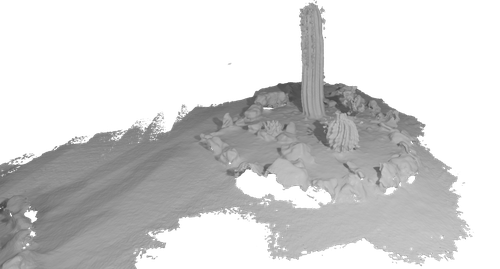}};
	\draw[red,thick,rounded corners] (3.2,2.8) rectangle (3.5,3.2);
	\draw[red,thick,rounded corners] (4.25,3.4) rectangle (4.95, 4.1);
	\end{tikzpicture} &
	\begin{tikzpicture}
	\node[anchor=south west,inner sep=0] at (0,0) {\includegraphics[width=\sz\textwidth]{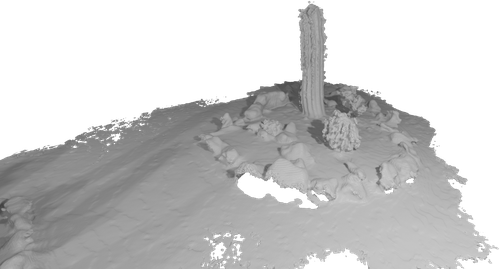}};
	\draw[red,thick,rounded corners] (3.2,2.8) rectangle (3.5,3.2);
	\draw[red,thick,rounded corners] (4.25,3.4) rectangle (4.95, 4.1);
	\end{tikzpicture} \\
	\multirow{1}{*}[85pt]{\rotatebox{90}{Cactus garden - top view}} & &
	\begin{tikzpicture}
	\node[anchor=south west,inner sep=0] at (0,0) {\includegraphics[width=\sz\textwidth]{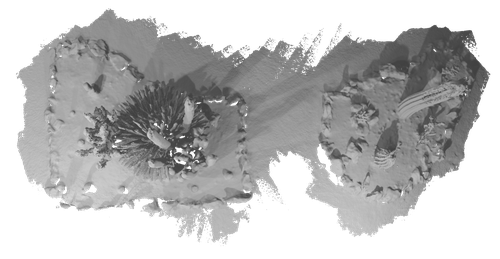}};
	\draw[red,thick,rounded corners] (3.75,0.8) rectangle (4.7,1.9);
	\end{tikzpicture} &
	\begin{tikzpicture}
	\node[anchor=south west,inner sep=0] at (0,0) {\includegraphics[width=\sz\textwidth1]{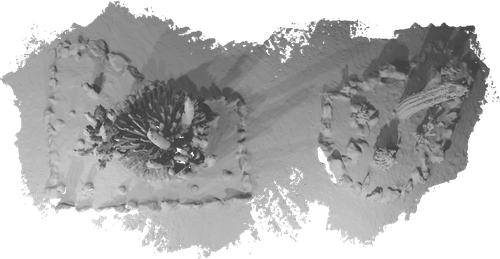}};
	\draw[red,thick,rounded corners] (3.75,0.8) rectangle (4.7,1.9);
	\end{tikzpicture} \\
	\multirow{1}{*}[70pt]{\rotatebox{90}{Lounge - top view}} & &
	\begin{tikzpicture}
	\node[anchor=south west,inner sep=0] at (0,0) {\includegraphics[width=\sz\textwidth]{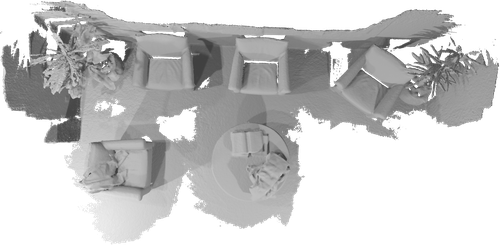}};
	\draw[red,thick,rounded corners] (2.2,1.5) rectangle (3.0,2.1);
	\end{tikzpicture} &
	\begin{tikzpicture}
	\node[anchor=south west,inner sep=0] at (0,0) {\includegraphics[width=\sz\textwidth]{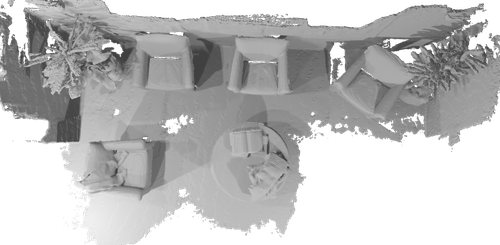}};
	\draw[red,thick,rounded corners] (2.2,1.5) rectangle (3.0,2.1);
	\end{tikzpicture} \\
	\multirow{1}{*}[90pt]{\rotatebox{90}{Lounge - Chair close-up}} & &
	\begin{tikzpicture}
	\node[anchor=south west,inner sep=0] at (0,0) {\includegraphics[width=\sz\textwidth]{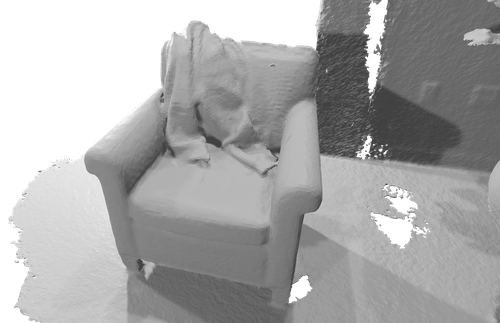}};
	\draw[red,thick,rounded corners] (3.8,3.6) rectangle (4.2,4);
	\draw[red,thick,rounded corners] (5.2,1.0) rectangle (6.3,4.85);
	\end{tikzpicture} &
	\begin{tikzpicture}
	\node[anchor=south west,inner sep=0] at (0,0) {\includegraphics[width=\sz\textwidth]{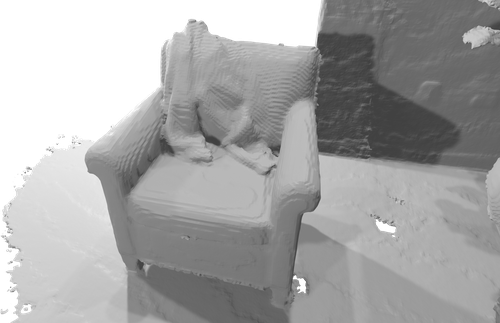}};
	\draw[red,thick,rounded corners] (3.8,3.6) rectangle (4.2,4);
	\draw[red,thick,rounded corners] (5.2,1.0) rectangle (6.3,4.85);
	\end{tikzpicture} \\
	\multirow{1}{*}[75pt]{\rotatebox{90}{Stonewall}} & &
	\begin{tikzpicture}
	\node[anchor=south west,inner sep=0] at (0,0) {\includegraphics[width=\sz\textwidth]{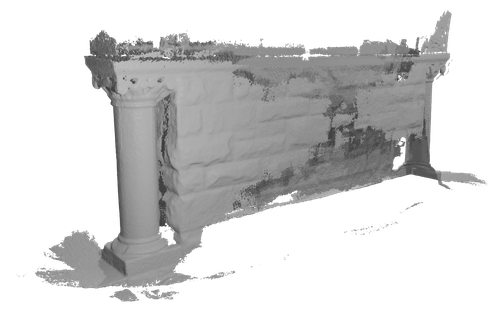}};
	\draw[red,thick,rounded corners] (3.3,1.6) rectangle (6.2,3.3);
	\end{tikzpicture} &
	\begin{tikzpicture}
	\node[anchor=south west,inner sep=0] at (0,0) {\includegraphics[width=\sz\textwidth]{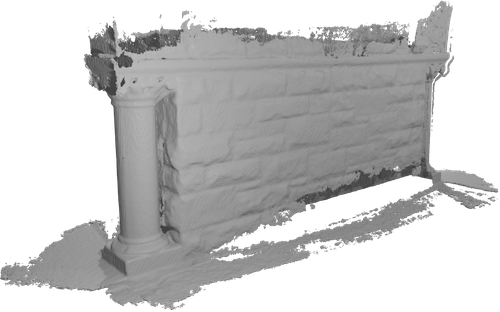}};
	\draw[red,thick,rounded corners] (3.3,1.6) rectangle (6.2,3.3);
	\end{tikzpicture} \\[-8pt]
	& & \hspace{2em}RoutedFusion~\cite{Weder-et-al-CVPR-2020} & \hspace{3em} Ours \\[-7pt]
	\end{tabular}
	\caption{\textbf{Additional results on Scene3D~\cite{Zhou-et-al-SIGGRAPH-2013}}. Our method reconstructs scenes with significantly higher completeness. This is due to the learned translation that can effectively discriminate between outliers and geometry. Furthermore, our method can filter large outlier blobs.}
	\label{fig:more-qualitative-scene3d}
\end{figure*}

In \figurename~\ref{fig:more-qualitative-scene3d} we show more results on the real-world Scene3D~\cite{Zhou-et-al-SIGGRAPH-2013} dataset. 
With this experiment, we demonstrate that our method generalizes well to real-world data and is able to fuse and reconstruct measurements of large-scale real-world scenes.

\section{Further Evaluation}
\subsection{Generalization from a Single Object}
In order to demonstrate the compactness and generalization performance of our network, we train it only on a single chair object from the ModelNet~\cite{Wu-et-al-CVPR-2015} dataset.
We augment the input depth maps with artificial noise of scale $0.01$.
We report the results in \tablename~\ref{tab:single-object-training} and show that our method trained on a single object achieves almost the same performance as our method trained on the full training set. 
Moreover, it outperforms the currently best performing method - RoutedFusion~\cite{Weder-et-al-CVPR-2020} - that is trained on the full training set.
\begin{table}[th]
	\centering
	\scriptsize
	\caption{Our method trained on the standard training split and on a single chair object only. The model trained on a single object is almost on par with our model trained on the full training set and outperforms the next best existing method trained on the full training set.}
	\label{tab:single-object-training}
	\setlength{\tabcolsep}{6pt}
	\begin{tabular}{lrrrr}
		\toprule
		\textbf{Method}  & \textbf{MSE}$\downarrow$  & \textbf{MAD}$\downarrow$   & \textbf{Acc.}$\uparrow$ & \textbf{IoU}$\uparrow$ \\
		& [e-05] & [e-02] & [\%] & [0,1] \\
		\midrule
		RoutedFusion~\cite{Weder-et-al-CVPR-2020} (full training set) & 6.79 & 0.56 & 94.44 & 0.821 \\
		Ours (full training set) & 4.84 & \textbf{0.42} & \textbf{96.30} & \textbf{0.874} \\
		Ours (single object)     & \textbf{3.94} & 0.44 & 94.51 & 0.848 \\
		\bottomrule
	\end{tabular}
\end{table}
This result indicates the applicability of our method to many real-world scenarios, where the sensor setup might change. 
In fact, only very little training data is required to retrain our method and achieve state-of-the-art reconstruction results.

\subsection{Loss Ablation}
We have also run an ablation study to evaluate the importance of the different terms in our loss function. 
In \figurename{}~\ref{fig:loss-ablation}, we show that the combination of all three loss terms yields best results.
The binary cross entropy is particularly useful to improve convergence in the beginning of the training as the network learns to predict a coarse shape that is further refined by the losses on the SDF as training progresses.

\clearpage

{\small
	\bibliographystyle{ieee_fullname}
	\bibliography{bibliography}
}

\end{document}